\documentclass{article} 
\usepackage[final]{colm2026_conference}

\usepackage{xcolor}
\usepackage{xspace}

\newcommand{\ours}{SynPro\xspace}

\definecolor{midnightgreen}{rgb}{0.0, 0.29, 0.33}

\usepackage{hyperref}
\definecolor{codegreen}{rgb}{0,0.6,0}
\definecolor{prevcitegreen}{rgb}{0.0, 0.42, 0.24} %
\definecolor{citegreen}{rgb}{0.0, 0.42, 0.24} 
\definecolor{codegray}{rgb}{0.5,0.5,0.5}
\definecolor{codepurple}{rgb}{0.58,0,0.82}
\definecolor{backcolour}{rgb}{0.95,0.95,0.92}

\definecolor{midnightblue}{rgb}{0.11, 0.11, 0.6} %
\hypersetup{
  colorlinks=true, 
  linkcolor=midnightblue,  
  citecolor=midnightblue,   
}

\usepackage{microtype}
\usepackage{amsmath}
\usepackage{hyperref}
\usepackage{url}
\usepackage{etoc}
\usepackage{booktabs}
\usepackage{wrapfig}
\usepackage{graphicx}
\usepackage{subcaption}
\usepackage{algorithm}
\usepackage{algpseudocode}
\usepackage{multicol}
\usepackage{titletoc}
\usepackage[table]{xcolor}
\usepackage[most]{tcolorbox}

\usepackage{lineno}

\definecolor{darkblue}{rgb}{0, 0, 0.5}
\hypersetup{colorlinks=true, citecolor=darkblue, linkcolor=darkblue, urlcolor=darkblue}

\title{Generating Pretraining Tokens from Organic Data for Data-Bound Scaling}

\author{Zichun Yu\textsuperscript{1}, Chenyan Xiong\textsuperscript{1,2} \\
  \textsuperscript{1}Language Technologies Institute, Carnegie Mellon University\quad
  \textsuperscript{2}Xlue \\
  \texttt{\{zichunyu,cx\}@andrew.cmu.edu}
}

\begin{document}

\ifcolmsubmission
\linenumbers
\fi

\maketitle

\begin{abstract}

  LLM pretraining is shifting from a compute-bound to a data-bound regime, where
  available human (organic) text falls far short of scaling demands.
  However, reaching the data-bound regime does not mean the model has fully utilized its
  organic corpus. In this paper, we introduce \textsc{\ours}, a synthetic data generation
  framework that
  helps LLMs more thoroughly learn from limited organic data.
  \textsc{\ours} applies two operations, rephrasing and reformatting, that present
  the same organic source in diverse forms to facilitate deeper learning without
  introducing external information.
  Both generators are optimized via reinforcement learning with quality, faithfulness,
  and data influence rewards, and are continuously updated as pretraining plateaus to
  target content the model has yet to absorb.
  We pretrain 400M, 1.1B, and 2B models with 10\% of their Chinchilla-optimal
  tokens (0.8B, 2.2B, and 4B) from DCLM-Baseline, reflecting a realistic data-bound regime in
  frontier pretraining.
  Our results reveal that organic data is significantly
  underutilized by standard repetition: \textsc{\ours} unlocks
  $3.4$--$5.2\times$ the effective tokens of repetition, even surpassing the
  non-data-bound oracle that trains
  on equivalent unique data at the 1.1B and 2B scales.
  Analyses confirm that faithful, model-aware synthesis sustains data-bound scaling
  without causing distribution collapse.
  We open-source our code at \url{https://github.com/cxcscmu/SynPro}.


\end{abstract}
\section{Introduction}

\begin{figure}[h!]
  \vspace{-0.2cm}
  \centering
  \begin{subfigure}[t]{0.46\textwidth}
    \centering
    \includegraphics[width=1.0\linewidth]{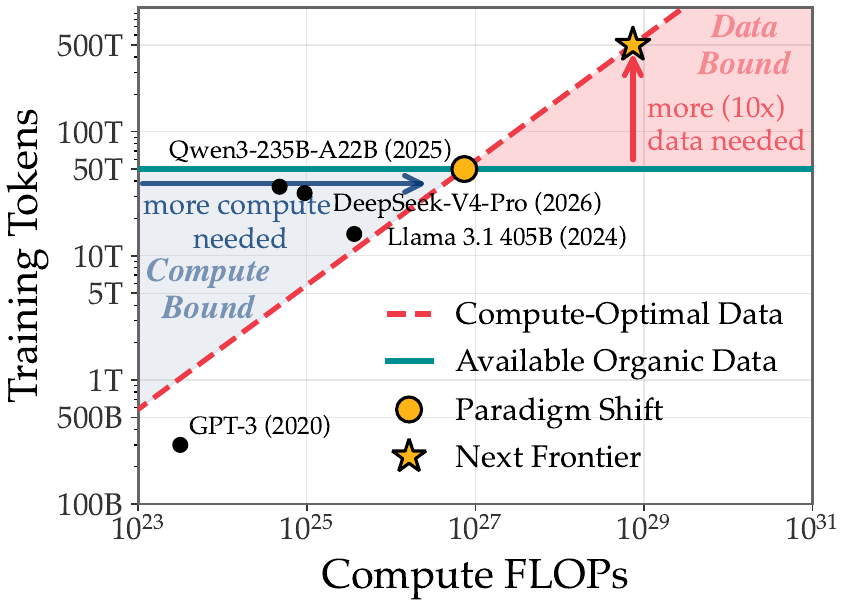}
    \caption{Paradigm shift in frontier model pretraining}
    \label{fig:paradigm-shift}
  \end{subfigure}
  \hfill
  \begin{subfigure}[t]{0.46\textwidth}
    \centering
    \includegraphics[width=1.0\linewidth]{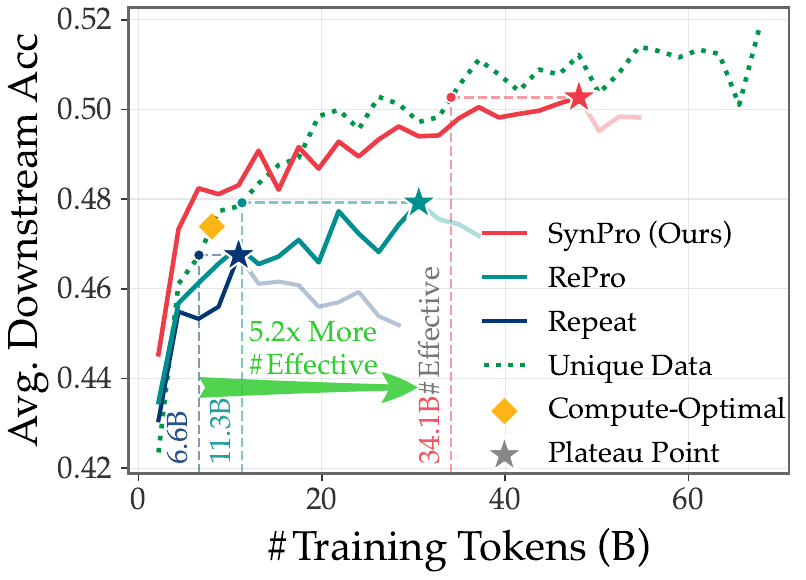}
    \caption{400M model \& 0.8B unique organic tokens}
    \label{fig:400m-curves}
  \end{subfigure}
  \caption{(a)~Paradigm shift in frontier pretraining from compute-bound to data-bound.
    (b)~Typical data-bound setup (400M model, 1/10 compute-optimal data); 1.1B in
  Figure~\ref{fig:1b-curves}.}
  \vspace{-0.3cm}
  \label{fig:scaling-curves}
\end{figure}




``We've achieved peak data and there'll be no more.''\footnote{Attributed to Ilya
Sutskever in public remarks discussing the limits of available pretraining data.}
As shown in Figure~\ref{fig:paradigm-shift}, frontier model pretraining
is undergoing a paradigm shift from compute-bound to data-bound scaling:
while the compute-optimal data requirement~\citep{Chinchilla} increases steadily with
surging compute, the growth of high-quality human (organic) text can no longer keep
pace~\citep{villalobos2022trends,maini2025beyondweb}.
Beyond the transition point, the next frontier of scaling demands an order of
magnitude more data than is currently available~\citep{villalobos2024will}.
In this data-bound regime, training with repeated passes over the available corpus
often yields diminishing returns and rapid saturation~\citep{muennighoff2023scaling}.


To further scale pretraining beyond the
organic data limit, synthetic data emerges as a practical
path~\citep{maini2024rephrasing,allal2024cosmopedia,maini2025beyondweb}.
However, unconstrained generation can lead to distribution
collapse~\citep{shumailov2024nature,dohmatob2025strong} or distill the generator's
parametric knowledge in ways that hurt generalization~\citep{chen2024diversity}.
At the same time, prior work~\citep{frank2023bridging,warstadt-etal-2023-findings} suggests that
effective learning may require much less data than current pretraining practice,
implying that available data may still be underutilized. These motivate a more
constrained use of synthetic data: \textit{Can we generate synthetic data grounded in
organic data to help LLMs learn more?}


In this paper, we introduce \textsc{\ours}, a synthetic data generation framework that
helps pretraining models learn more thoroughly from limited organic data.
\textsc{\ours} generates data through two operations that facilitate model learning:
\textit{rephrasing}~\citep{yu2025repro}, which introduces
lexical and syntactic
diversity while
preserving core semantics, and \textit{reformatting}~\citep{su2025nemotroncc}, which converts
source content into task-oriented forms.
Both operations are optimized via reinforcement learning with
a \textit{quality} reward that ensures coherent text,
a \textit{faithfulness} reward that grounds outputs in the source document,
and a \textit{data influence} reward~\citep{yu2024mates} that steers
generation toward content the current pretraining model has yet to absorb.
\textsc{\ours} continuously updates the generator to produce informative
yet grounded data that helps the model continue improving.

We pretrain 400M, 1.1B, and 2B models with 10\% of their Chinchilla-optimal tokens
from the English corpus DCLM-Baseline~\citep{li2024datacomp}, reflecting a realistic
data-bound regime in frontier pretraining.
As shown in Figure~\ref{fig:400m-curves}, our results reveal that organic pretraining
data is significantly underutilized by
standard repetition: \textsc{\ours} unlocks $5.2\times$ the \textit{effective
tokens} (equivalent unique data yielding the same performance) of simple repetition
and $3.0\times$ those of RePro~\citep{yu2025repro}, the state-of-the-art web rephrasing
baseline.
\textsc{\ours} stays above repetition at every point of the curve, while repetition
quickly degrades once it plateaus.
At the 1.1B and 2B scales, \textsc{\ours} even surpasses training on the same amount of
unique organic data, demonstrating that faithful synthesis can unlock significantly
more value from limited data for LLM pretraining.

To better understand why these gains arise, we first show that our synthetic data
preserves both pointwise and distributional properties of the organic corpus rather than
collapsing toward a narrow mode, confirming the value of grounded generation for
sustained pretraining gains.
Furthermore, our generator adaptively shifts its output toward content
the current model has yet to absorb, producing more informative data throughout training
where static approaches decay.
These results highlight that faithful, model-aware synthesis can sustain data-bound scaling
without causing distribution collapse or relying on distillation.

We summarize our contributions as follows:
\begin{enumerate}
  \item We propose \textsc{\ours}, a model-aware synthetic data generation
    framework that helps pretraining models more thoroughly utilize a limited organic corpus.
  \item We systematically define and study the data-bound regime, where \textsc{\ours} achieves
    up to $5.2\times$ the effective tokens over repetition, approaching the unique data oracle.
  \item \textsc{\ours} reveals that organic pretraining data is underutilized rather
    than exhausted, and faithful synthesis can unlock more value from it without
    distribution collapse.
\end{enumerate}
\section{Related work}
\vspace{-0.2cm}


\paragraph{LLM scaling and data wall.}
Progress in large language models has been driven by jointly scaling parameters,
computation, and training data~\citep{kaplan2020scaling,Chinchilla}.
Compute is no longer the primary bottleneck thanks to hardware improvements and
architectural innovations~\citep{shazeer2017outrageously,liu2024deepseekv3};
instead, projections suggest that publicly available human-written text will be
insufficient to sustain current scaling
trajectories~\citep{villalobos2022trends,villalobos2024will,shen2025will}.
When the available data falls well below the compute-optimal
requirement~\citep{Chinchilla}, practitioners often resort to repeated passes over
the same corpus, which yields diminishing gains after only a
few (typically 4) epochs~\citep{muennighoff2023scaling}. This phenomenon, known as the
data wall, leads to a plateau in performance despite increased training time.

\paragraph{Synthetic data for pretraining.}
Generating synthetic text is a natural strategy to augment a limited organic
corpus~\citep{havrilla2024surveying}.
Effective methods include document-level
paraphrasing~\citep{maini2024rephrasing}, guided
rewriting~\citep{nguyen2025rewire}, and textbook-style
generation~\citep{li2023textbooks,allal2024cosmopedia,hao2025reformulation}, which
improve pretraining data volume and
quality~\citep{su2025nemotroncc,abdin2024phi4,maini2025beyondweb}.
We follow the prompt design of Nemotron-CC for reformatting, but
\textit{train the generator with explicit rewards} instead of relying on static prompts.
However, unconstrained synthesis risks model
collapse~\citep{dohmatob2024tale,dohmatob2025strong}: successive training on
synthetic data erodes the tail distribution~\citep{shumailov2024nature}, and negative
effects can propagate to post-training~\citep{chen2024diversity}.
These findings highlight that grounding synthetic data in the source content is essential.
To keep such faithfulness, ProX~\citep{zhou2024prox} and RefineX~\citep{bi2025refinex}
restrict editing to conservative operations such as deletion and normalization,
while RePro~\citep{yu2025repro} optimizes a rephraser via quality and faithfulness
rewards to produce high-quality yet grounded data.
Unlike RePro that recycles low-quality web data, we target the \textit{data-bound
scaling} regime, which motivates a \textit{model-aware influence reward} to keep mining
new signal from the same organic corpus.

\paragraph{Model-aware data curation.}
A line of work on curriculum learning~\citep{zhang2026beyond} tailors data
strategies to the model's evolving needs.
On the selection side, DsDm~\citep{engstrom2024dsdm}, MATES~\citep{yu2024mates} and
GREATS~\citep{wang2024greats} leverage data influence~\citep{koh2017understanding} to
prioritize informative
samples, while CLIMB~\citep{diao2025nemotronclimb} tunes domain
proportions through iterative search guided by a proxy performance predictor.
On the generation side, Montessori-Instruct~\citep{li2025montessori} trains a teacher
to produce high-influence instruction-tuning examples for a target student and both
models are updated in tandem. These advances demonstrate the strong potential of model-aware
optimization to enhance data curation.
Our work extends such model-aware optimization to \textit{generating pretraining
tokens} for data-bound scaling.

\section{Method}

\begin{figure*}
  \centering
  \includegraphics[width=0.99\linewidth]{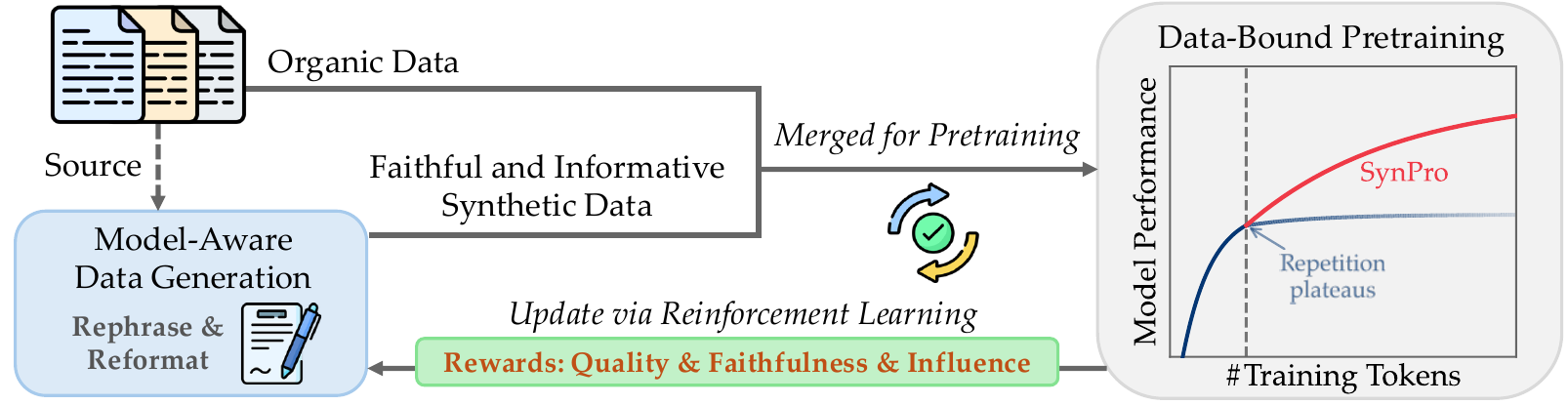}
  \caption{Overview of \textsc{SynPro}. We train generators to provide
    faithful and informative synthetic data from organic source, enabling sustained
  improvement for data-bound scaling.}
  \label{fig:pipeline}
  \vspace{-0.5cm}
\end{figure*}

This section presents \textsc{\ours} (Figure~\ref{fig:pipeline}), an effective and
faithful synthetic data generation
framework that helps the pretraining model better utilize a limited organic corpus.

\subsection{Data-bound scaling regime}

We assume access to an organic corpus $\mathcal{D}_{\text{org}}$, comprising all
web-sourced data available for pretraining.
Given a compute budget $C$, the compute-optimal data requirement is
$D^{*}(C)$ tokens~\citep{Chinchilla}. We define the \textit{available data ratio} $\alpha$ as:
\begin{align}
  \alpha = \frac{|\mathcal{D}_{\text{org}}|}{D^{*}(C)}.
\end{align}
In the early stages of LLM development, data was abundant relative to compute $C$,
so $\alpha > 1$ and scaling was \textit{compute-bound}.
As illustrated in Figure~\ref{fig:paradigm-shift}, frontier pretraining is now
shifting to a \textit{data-bound} regime
where publicly available text is approaching exhaustion~\citep{villalobos2024will}.
Meanwhile, \cite{Chinchilla} predict that the next frontier requires
roughly $100\times$ more compute and $10\times$ more data, which places $\alpha$ at
around \textbf{10\%}, meaning the available organic data covers
only a fraction of compute-optimal requirements.
The standard practice in this regime is to repeatedly train the language model
$\mathcal{M}$ on $\mathcal{D}_{\text{org}}$, but this yields diminishing returns
after only a few epochs~\citep{muennighoff2023scaling}.

\subsection{Model-aware synthetic data generation}
\label{sec:method-framework}

To overcome data limitations, \textsc{\ours} enables a better utilization of the limited
organic corpus by
synthesizing grounded and informative data $\mathcal{D}_{\text{syn}}$ across three
repeating stages: (1) LM pretraining,
(2) generation policy update, and (3) generation of new synthetic data.

To start the process, we apply an initial generation policy $\pi_0$ to each organic sample
$x \in \mathcal{D}_{\text{org}}$, conditioned on a prompt $p$, to produce
an initial synthetic corpus and training set:
\begin{align}
  \mathcal{D}_{\text{syn}}^{0}
  &= \{\pi_0(p, x) \mid x \in \mathcal{D}_{\text{org}}\}, \\
  \mathcal{D}_{\text{train}}
  &= \mathcal{D}_{\text{org}} \cup \mathcal{D}_{\text{syn}}^{0}.
\end{align}
\paragraph{Stage 1: LM pretraining.}
At each iteration $i$, we continue pretraining on $\mathcal{D}_{\text{train}}$ from the
previous checkpoint $\mathcal{M}_{i-1}^{*}$ ($\mathcal{M}_{0}^{*}$ is randomly
initialized) until the reference loss
$\mathcal{L}(\mathcal{D}_{\text{ref}} \mid \mathcal{M})$ saturates. Here, the reference set
$\mathcal{D}_{\text{ref}}$ serves as a proxy for the model’s
generalization and does not overlap with the downstream evaluation.
We define saturation as the point where the reference loss fails to improve over the best
of the previous \textit{two epochs}, formally:
\begin{align}
  &\mathcal{M}_i = \mathcal{M}_{i-1}^{*},~\mathcal{L}_0 =
  \mathcal{L}(\mathcal{D}_{\text{ref}} \mid \mathcal{M}_{i-1}^{*}), \\
  &\text{For } t = 1, 2, \ldots: \\
  &\quad\quad \mathcal{M}_i^{'} \leftarrow \mathcal{A}(\mathcal{M}_i,\,
  \mathcal{D}_{\text{train}}),~\mathcal{L}_t = \mathcal{L}(\mathcal{D}_{\text{ref}} \mid
  \mathcal{M}_i^{'}), \\
  &\quad\quad \text{If } t > 1 \text{ and } \mathcal{L}_t \geq \min(\mathcal{L}_{t-1},\,
  \mathcal{L}_{t-2})\text{:}~\textbf{break}\\
  &\quad\quad \mathcal{M}_i = \mathcal{M}_i^{'}, \\
  &\mathcal{M}_i^{*} = \mathcal{M}_i,
\end{align}
where $\mathcal{A}(\mathcal{M},\, \mathcal{D}_{\text{train}})$ denotes one epoch of
training on $\mathcal{D}_{\text{train}}$ starting from checkpoint $\mathcal{M}$.

\paragraph{Stage 2: Policy update.}
When model pretraining saturates, we update the generation policy to produce synthetic
data that is more informative for the current saturated model $\mathcal{M}_i^{*}$.
The policy is optimized under a composite reward:
\begin{align}
  r_i(x, \tilde{x})
  &= \lambda_{\text{quality}}\,r_{\text{quality}}(\tilde{x})
  + \lambda_{\text{faithful}}\,r_{\text{faithful}}(x, \tilde{x})
  + \lambda_{\text{influence}}\,r_{\text{influence}}(\tilde{x} \mid \mathcal{M}_i^{*}),
\end{align}
where $\lambda_{\text{quality}}$, $\lambda_{\text{faithful}}$, and
$\lambda_{\text{influence}}$ control the relative weight of each reward component.
The quality, faithfulness, and data influence rewards ensure that generated synthetic
data is written in high-quality language, grounded in the source document, and targeted
at what the current model has yet to learn (detailed in \S\ref{sec:method-operations}).
We update the policy with standard reinforcement learning to maximize the expected reward:
\begin{align}
  \pi_{i}
  &= \arg\max_{\pi}\;
  \mathbb{E}_{x\sim\mathcal{D}_{\text{org}},\,\tilde{x}\sim\pi(\cdot\mid p,x)}
  \!\left[r_i(x, \tilde{x})\right].
\end{align}
\paragraph{Stage 3: Generation of new synthetic data.}
The updated policy $\pi_i$ generates a fresh set of synthetic data from the organic
corpus, which is appended to the training set:
\begin{align}
  \mathcal{D}_{\text{syn}}^{i}
  = \{\pi_i(p, x) \mid x \in \mathcal{D}_{\text{org}}\},~\mathcal{D}_{\text{train}}
  \leftarrow \mathcal{D}_{\text{train}} \cup \mathcal{D}_{\text{syn}}^{i}.
\end{align}
We then return to Stage~1 until the reference loss fails to improve in the next
iteration, i.e., $\mathcal{L}(\mathcal{D}_{\text{ref}} \mid \mathcal{M}_{i}^{*}) \geq
\mathcal{L}(\mathcal{D}_{\text{ref}} \mid
\mathcal{M}_{i-1}^{*})$. Together, \textsc{\ours} continuously
provides effective synthetic data for data-bound pretraining.
Algorithm~\ref{alg:synpro} summarizes the entire pipeline.

\subsection{Synthetic data operations and reward design}
\label{sec:method-operations}

\textsc{\ours} employs two complementary operations (prompts) to synthesize helpful data from
$\mathcal{D}_{\text{org}}$, \textit{rephrasing} and \textit{reformatting}.
\textit{Rephrasing} rewrites each source document to diversify surface form (word
choice, grammar fix, clause ordering) while preserving the core
semantics~\citep{yu2025repro}.
\textit{Reformatting} transforms each document into a task-oriented form, such as a
comparative analysis, a knowledge highlight, or a reasoning
trace, allowing one source document to yield multiple distinct yet
grounded outputs~\citep{su2025nemotroncc}.
Table~\ref{tab:operation-examples} shows an example of each operation.

\begin{table}[t]
  \centering
  \renewcommand{\arraystretch}{1.2}
  \setlength{\tabcolsep}{5pt}
  \small
  \begin{tabular}{l|p{0.40\textwidth}|p{0.40\textwidth}}
    \toprule
    \textbf{Operation} & \textbf{Organic source (excerpt)} &
    \textbf{Synthetic output (excerpt)} \\
    \midrule
    Rephrasing & ``The acquisition of embryogenic competence has been achieved on a
    few occasions and with low induction rates
    \textcolor{red!70!black}{2,3-Butanediol [2,3,12]}.'' &
    ``Obtaining embryogenic competence has been achieved infrequently and with limited
    induction rates.'' \\
    \hline
    Reformatting & ``The Delphi method was developed in the United States in the 1950s
    and 1960s in the military domain.'' &
    ``\textit{Question:} What is the origin of the Delphi method?
    \textit{Answer:} Military in the United States in the 1950s and 1960s.'' \\
    \bottomrule
  \end{tabular}
  \caption{Examples of the two operations. Rephrasing preserves the semantics while
    polishing the surface form, including removing the noisy spans injected by web
    scraping (marked in red); reformatting restructures the same source content.
  Full cases are in Appendix~\ref{sec:app-cases}.}
  \label{tab:operation-examples}
\end{table}

Both operations share the quality and data influence rewards, which apply uniformly
regardless of the output format; faithfulness is defined differently for each due to
their distinct structural relationships to the source. We detail the reward design below.

\paragraph{Quality ($r_{\text{quality}}$).}
We adopt DataMan~\citep{peng2025dataman}, a tuned small LM that evaluates text across 13
quality criteria (e.g., coherence, topic focus) and gives an overall score:
\begin{align}
  r_{\text{quality}}(\tilde{x}) = \text{DataMan}(\tilde{x}).
\end{align}
This reward ensures the generator produces high-quality text that is well-formed,
coherent, and informative, which is crucial for effective pretraining.

\paragraph{Data influence ($r_{\text{influence}}$).}
Following MATES~\citep{yu2024mates} and Forward-INF~\citep{ko2024mirrored}, we efficiently compute
data influence as the loss reduction on a synthetic sample $\tilde{x}$ after the current
model is updated on the reference set $\mathcal{D}_{\text{ref}}$:
\begin{align}
  r_{\text{influence}}(\tilde{x} \mid \mathcal{M}_i^{*}) =
  \mathcal{L}(\tilde{x} \mid \mathcal{M}_i^{*}) -
  \mathcal{L}(\tilde{x} \mid \mathcal{A}(\mathcal{M}_i^{*}, \mathcal{D}_{\text{ref}})),
  \label{eq:influence}
\end{align}
where $\mathcal{A}(\mathcal{M}_i^{*}, \mathcal{D}_{\text{ref}})$ denotes training on
$\mathcal{D}_{\text{ref}}$. This reverses the oracle computation, so scoring each
sample requires only \textit{two forward passes} instead of a per-sample backward
pass, with a detailed derivation provided in Appendix~\ref{sec:app-influence}.
This reward steers the generator toward outputs useful to the current model, making the
synthetic data model-aware.

\paragraph{Faithfulness ($r_{\text{faithful}}$).}
For rephrasing, following RePro~\citep{yu2025repro}, we combine three binary rewards:
semantic similarity via BERTScore~\citep{zhang2020bertscore}, structural preservation via
LLM-as-a-judge, and a length constraint to penalize free-form generation:
\begin{align}
  r_{\text{faithful}}(x, \tilde{x})
  &= \mathbf{1}[\text{BERTScore}(x, \tilde{x}) \ge \tau_{\text{sem}}]
  \cdot \mathbf{1}[\text{Structure}(x, \tilde{x})]
  \cdot \mathbf{1}\!\left[\frac{\text{Len}(\tilde{x})}{\text{Len}(x)} \le
  \tau_{\text{len}}\right].
\end{align}
For reformatting, where surface form changes substantially, we instead train a small reward
model distilled from an LLM to classify whether the output is faithful to the source,
yielding a binary reward $r_{\text{faithful}}(x, \tilde{x}) \in \{0, 1\}$. The
faithfulness reward ensures synthetic data reflect the source document rather than the
generator's distilled knowledge.


\section{Experimental setup}
\label{sec:setup}
\vspace{-0.1in}

\paragraph{Pretraining model and data.}
We pretrain decoder-only Transformers~\citep{vaswani2017attention} from scratch at three
scales: a 1.1B model uses the OLMo2-1B~\citep{walsh2025olmo} architecture, together with a
400M scaled-down variant and a 2B scaled-up variant.
We randomly sample from
DCLM-Baseline~\citep{li2024datacomp} as the organic corpus $\mathcal{D}_{\text{org}}$, the
state-of-the-art open-source pretraining dataset.
The compute-optimal data requirements are 8B/22B/40B for the 400M/1.1B/2B
model~\citep{Chinchilla}.
To study different degrees of data limitation, we set
$\alpha \in \{5\%, 10\%, 15\%\}$ for the 400M model
(corresponding to 0.4B, 0.8B, and 1.2B organic tokens) and
$\alpha = 10\%$ for the 1.1B and 2B models (2.2B and 4B organic tokens, respectively).
10\% reflects a typical bottleneck in frontier
pretraining based on the current scaling trend~\citep{villalobos2024will}.
We use warmup and stable phases of the WSD scheduler~\citep{hu2024minicpm}. More details
are provided in
Appendix~\ref{sec:app-setup}.

\paragraph{Baselines.}
We compare \textsc{\ours} against four baselines:
\begin{enumerate}
  \item \textit{Repeat}: repeatedly training on the full organic data
    $\mathcal{D}_{\text{org}}$ until saturation.
  \item \textit{QBSR} (Quality-Based Selective Repetition)~\citep{luo2025pcmind}: a static
    curriculum that, after full repetition saturates, continues training by repeating
    only the top 80\%,
    60\%, 40\%, and 20\% of data ranked by quality scores (we choose DataMan here).
  \item \textit{MATES}~\citep{yu2024mates}: an adaptive curriculum that, after full
    repetition saturates, selects data with positive influence for the next
    epoch and iterates until saturation.
  \item \textit{RePro}~\citep{yu2025repro}: augmenting the organic corpus with
    rephrased data generated by OLMo2-1B-Instruct~\citep{walsh2025olmo} trained
    with quality and faithfulness rewards, which has been shown to outperform other
    web rephrasing methods such as WRAP~\citep{maini2024rephrasing} and
    ReWire~\citep{nguyen2025rewire}.
\end{enumerate}

We also report \textit{Unique Data}, which trains on unique organic data from
DCLM-Baseline, as a non-data-bound oracle. This comparison is not apples-to-apples to our method.

\paragraph{Evaluation.}
Following~\cite{walsh2025olmo}, we report \textit{zero-shot accuracy} on 9 downstream tasks:
ARC-Easy, ARC-Challenge~\citep{clark2018arc},
SciQ~\citep{welbl2018sciq},
OpenBookQA~\citep{mihaylov2018obqa},
HellaSwag~\citep{zellers2019hellaswag},
PIQA~\citep{bisk2020piqa},
WinoGrande~\citep{sakaguchi2021winogrande},
CommonsenseQA~\citep{talmor2019commonsenseqa},
and SIQA~\citep{sap2019siqa}.
These tasks provide a comprehensive assessment of commonsense reasoning, language
understanding, and knowledge.
We further define \textit{effective
tokens} as the amount of unique organic data that, trained for a single epoch, yields
the same performance, to quantify data efficiency. We also report \textit{recovery ratio},
the fraction of the performance gap from Repeat to Unique Data:
\begin{align}
  \text{Recovery Ratio} = \frac{\text{Method Acc} - \text{Repeat Acc}}
  {\text{Unique Acc} - \text{Repeat Acc}}.
  \label{eq:recovery-ratio}
\end{align}

\paragraph{Implementation details.}
We use FLAN~\citep{wei2022flan} as the held-out reference set
$\mathcal{D}_{\text{ref}}$ for computing data influence following~\cite{yu2025group}.
We initialize both generators with OLMo2-1B-Instruct and train them with quality and
faithfulness rewards to serve as $\pi_0$.
\begin{table*}[t]
  \centering
  \renewcommand{\arraystretch}{1.3}
  \setlength{\tabcolsep}{3pt}
  \resizebox{1.0\textwidth}{!}{%
    \begin{tabular}{l|cc|cccc|cc|ccc|cc}
      \toprule
      & \textbf{\#Training} & \textbf{\#Effective} &
      \multicolumn{4}{c|}{\textit{Commonsense Reasoning}} &
      \multicolumn{2}{c|}{\textit{Language Understanding}} &
      \multicolumn{3}{c|}{\textit{World Knowledge}} & & \textbf{Rec.} \\
      \textbf{Method} & \textbf{Tokens} & \textbf{Tokens} &
      \textbf{CSQA} & \textbf{OBQA} & \textbf{PIQA} & \textbf{SIQA} &
      \textbf{HellaSwag} & \textbf{WinoG} &
      \textbf{ARC-e} & \textbf{ARC-c} & \textbf{SciQ} &
      \textbf{Avg} & \textbf{Ratio} \\
      \midrule
      \multicolumn{14}{l}{\textbf{400M model, 8B Chinchilla-optimal tokens,
      $\alpha{=}5\%$ (0.4B available organic tokens)}} \\
      \midrule
      \textit{Unique}      & \textit{13.1B} & \textit{13.1B} & \textit{0.3415} &
      \textit{0.3340} & \textit{0.6480} & \textit{0.4145} & \textit{0.3731} &
      \textit{0.5130} & \textit{0.6070} & \textit{0.2876} & \textit{0.8330} &
      \textit{0.4835} & \textit{100\%} \\
      \hline
      Repeat              & 5.5B & 2.9B$_{~{1.0\times}}$ & 0.3006 & 0.2920 & 0.6045 &
      0.4110 & 0.3193 &
      0.5083 & 0.5088 & 0.2575 & 0.7230 & 0.4361 & 0\% \\
      QBSR               & 6.3B & 3.2B$_{~\textcolor{red}{1.1\times}}$ & 0.3071 & 0.3040
      & 0.6251 & \underline{0.4120} & 0.3225 & 0.5122
      & 0.5386 & 0.2341 & 0.7210 & 0.4418 & 12\% \\
      MATES               & 6.0B & 3.0B$_{~\textcolor{red}{1.0\times}}$ &
      \underline{0.3112} & 0.3060
      & 0.5990 & 0.4033 & 0.3209 & 0.5193
      & 0.5105 & \underline{0.2742} & 0.7010 & 0.4384 & 5\% \\
      RePro     & 7.6B & \underline{4.0B}$_{~\textcolor{red}{1.4\times}}$ & 0.3055 &
      \textbf{0.3200} & \underline{0.6322} & 0.4023 & \underline{0.3384} &
      \textbf{0.5320} & \underline{0.5474}
      & 0.2709 & \underline{0.7510} & \underline{0.4555} & \underline{41\%} \\
      \rowcolor{blue!10}
      \textsc{\ours} & 13.1B & \textbf{9.9B}$_{~\textcolor{red}{\textbf{3.4}\times}}$ &
      \textbf{0.3604} & \underline{0.3160} & \textbf{0.6393} &
      \textbf{0.4156} & \textbf{0.3485} & \underline{0.5249} & \textbf{0.6053} &
      \textbf{0.2843} & \textbf{0.8070} & \textbf{0.4779} & \textbf{88\%} \\
      \midrule
      \multicolumn{14}{l}{\textbf{400M model, 8B Chinchilla-optimal tokens,
      $\alpha{=}10\%$ (0.8B available organic tokens)}} \\
      \midrule
      \textit{Unique}      & \textit{48.0B} & \textit{48.0B} & \textit{0.3686} &
      \textit{0.3360} & \textit{0.6790} & \textit{0.4253} & \textit{0.4303} &
      \textit{0.5399} & \textit{0.6579} & \textit{0.3211} & \textit{0.8510} &
      \textit{0.5121} & \textit{100\%} \\
      \hline
      Repeat               & 10.9B & 6.6B$_{~{1.0\times}}$ & 0.3358 & 0.3140 & 0.6442 &
      0.4135 & 0.3519 &
      \underline{0.5304} & 0.5579 & 0.2676 & 0.7920 & 0.4675 & 0\% \\
      QBSR               & 12.5B & 6.9B$_{~\textcolor{red}{1.0\times}}$ & 0.3342 & 0.3160
      & 0.6333 & 0.4140 &
      0.3533 & 0.5209 & 0.5754 & \underline{0.2742} & 0.8000 & 0.4690 & 3\% \\
      MATES               & 12.1B & 7.5B$_{~\textcolor{red}{1.1\times}}$ &
      \underline{0.3415} & \underline{0.3200} & 0.6436 & 0.4150 & 0.3602 & 0.5012 &
      0.5684 & 0.2709 & \underline{0.8220} & 0.4714 & 9\% \\
      RePro     & 30.6B & \underline{11.3B}$_{~\textcolor{red}{1.7\times}}$ & 0.3227 &
      \textbf{0.3420} & \underline{0.6540} &
      \underline{0.4197} & \underline{0.3783} & \textbf{0.5312} & \underline{0.5947} &
      \underline{0.2742} & 0.7960 & \underline{0.4792} & \underline{26\%} \\
      \rowcolor{blue!10}
      \textsc{\ours}       & 48.0B &
      \textbf{34.1B}$_{~\textcolor{red}{\textbf{5.2}\times}}$ & \textbf{0.3882} & \textbf{0.3420} &
      \textbf{0.6649} & \textbf{0.4284} & \textbf{0.3868} & 0.5201 & \textbf{0.6123} &
      \textbf{0.3278} & \textbf{0.8540} & \textbf{0.5027} & \textbf{79\%} \\
      \midrule
      \multicolumn{14}{l}{\textbf{400M model, 8B Chinchilla-optimal tokens,
      $\alpha{=}15\%$ (1.2B available organic tokens)}} \\
      \midrule
      \textit{Unique}         & \textit{61.2B} & \textit{61.2B} & \textit{0.3636} &
      \textit{0.3600} & \textit{0.6817} & \textit{0.4284} & \textit{0.4327} &
      \textit{0.5351} & \textit{0.6351} & \textit{0.3211} & \textit{0.8620} &
      \textit{0.5133} & \textit{100\%} \\
      \hline
      Repeat               & 17.5B & 7.9B$_{~{1.0\times}}$ & 0.3309 & \underline{0.3360} & 0.6420 &
      0.4222 & 0.3750 &
      0.5249 & 0.5456 & 0.2876 & 0.7960 & 0.4734 & 0\% \\
      QBSR               & 19.9B & 8.6B$_{~\textcolor{red}{1.1\times}}$ & 0.3268 & 0.3120
      & 0.6474 & 0.4232 & 0.3806 & 0.5185
      & 0.5825 & 0.3010 & 0.7980 & 0.4767 & 8\% \\
      MATES              & 19.7B & 8.7B$_{~\textcolor{red}{1.1\times}}$ & 0.3366 & 0.3260
      & 0.6453 & 0.4263 & 0.3766 &
      \underline{0.5288}
      & 0.5667 & 0.2910 & 0.7970 & 0.4771 & 9\% \\
      RePro     & 39.3B & \underline{17.9B}$_{~\textcolor{red}{2.3\times}}$ & \underline{0.3497} &
      0.3260 & \underline{0.6561} &
      \underline{0.4273} & \underline{0.3998} & 0.5099 & \underline{0.6263} &
      \underline{0.3077} & \underline{0.8170} & \underline{0.4911} & \underline{44\%} \\
      \rowcolor{blue!10}
      \textsc{\ours}       & 61.2B &
      \textbf{67.6B}$_{~\textcolor{red}{\textbf{8.6}\times}}$ & \textbf{0.4054} & \textbf{0.3500} &
      \textbf{0.6687} & \textbf{0.4365} & \textbf{0.4167} & \textbf{0.5304} & \textbf{0.6596} &
      \textbf{0.3211} & \textbf{0.8690} & \textbf{0.5175} & \textbf{111\%} \\
      \midrule
      \multicolumn{14}{l}{\textbf{1.1B model, 22B Chinchilla-optimal tokens,
      $\alpha{=}10\%$ (2.2B available organic tokens)}} \\
      \midrule
      \textit{Unique}         & \textit{56.8B} & \textit{56.8B} & \textit{0.4062} &
      \textit{0.3720} & \textit{0.7095} & \textit{0.4427} & \textit{0.5261} &
      \textit{0.5564} & \textit{0.7193} & \textit{0.3813} & \textit{0.8910} &
      \textit{0.5561} & \textit{100\%} \\
      \hline
      Repeat               & 39.3B & 15.5B$_{~{1.0\times}}$ & 0.3694 &
      0.3420 & \underline{0.6926} &
      0.4304 & 0.4589 & 0.5304 & 0.6526 & \underline{0.3344} & 0.8450 & 0.5173 & 0\% \\
      QBSR               & 43.7B & 13.7B$_{~\textcolor{red}{0.9\times}}$ & 0.3784 & 0.3340 &
      0.6828 & 0.4268 & 0.4450 & 0.5233 & 0.6561 & 0.3311 & 0.8590 & 0.5152 & $-$5\% \\
      MATES               & 42.6B & 15.9B$_{~\textcolor{red}{1.0\times}}$ & 0.3669 &
      \underline{0.3500} & 0.6910 & 0.4340 & 0.4577 &
      0.5367 & 0.6509 & 0.3278 & 0.8520 & 0.5185 & 3\% \\
      RePro     & 45.9B & \underline{21.7B}$_{~\textcolor{red}{1.4\times}}$ & \underline{0.4029} &
      0.3460 &
      \textbf{0.6942} & \underline{0.4371} & \underline{0.4893} & \textbf{0.5446} &
      \underline{0.6649} & 0.3311 & \underline{0.8770} & \underline{0.5319} & \underline{38\%} \\
      \rowcolor{blue!10}
      \textsc{\ours}       & 56.8B &
      \textbf{57.4B}$_{~\textcolor{red}{\textbf{3.7}\times}}$ & \textbf{0.4586} & \textbf{0.3520} &
      0.6910 & \textbf{0.4678} & \textbf{0.4917} & \underline{0.5391} &
      \textbf{0.7018} & \textbf{0.4147} & \textbf{0.9090} & \textbf{0.5584} & \textbf{106\%} \\
      \midrule
      \multicolumn{14}{l}{\textbf{2B model, 40B Chinchilla-optimal tokens,
      $\alpha{=}10\%$ (4B available organic tokens)}} \\
      \midrule
      \textit{Unique}         & \textit{69.9B} & \textit{69.9B} & \textit{0.4169} &
      \textit{0.3940} & \textit{0.7345} & \textit{0.4468} & \textit{0.5648} &
      \textit{0.5912} & \textit{0.7368} & \textit{0.4080} & \textit{0.9070} &
      \textit{0.5778} & \textit{100\%} \\
      \hline
      Repeat               & 52.4B & \underline{21.8B}$_{~{1.0\times}}$ &
      \underline{0.3980} & \underline{0.3700} &
      \underline{0.7013} & \underline{0.4268} & \underline{0.5208} & \underline{0.5430} &
      \underline{0.6947} & \underline{0.3746} & \underline{0.8730} & \underline{0.5447} &
      \underline{0\%} \\
      \rowcolor{blue!10}
      \textsc{\ours}       & 69.9B &
      \textbf{73.9B}$_{~\textcolor{red}{\textbf{3.4}\times}}$ & \textbf{0.4726} &
      \textbf{0.4160} &
      \textbf{0.7111} & \textbf{0.4570} & \textbf{0.5391} & \textbf{0.5699} &
      \textbf{0.7439} & \textbf{0.4415} & \textbf{0.9220} & \textbf{0.5859} & \textbf{124\%} \\
      \bottomrule
    \end{tabular}
  }
  \caption{Data-bound pretraining results.
    Every method is trained to its own saturation point (\S\ref{sec:method-framework}),
    so \textbf{\#Training Tokens} differ across methods.
    \textit{Unique} denotes the oracle performance obtained by training on all unique
    tokens, i.e., non-data-bound regime.
    \textbf{Bold} and \underline{underline} indicate the best and second-best results
    among methods with the same available organic corpus, so the \textit{Unique} oracle
  is excluded.}
  \label{tab:main}
  \vspace{-0.5cm}
\end{table*}

For rephrasing, we prompt Qwen3-1.7B~\citep{yang2025qwen3} for structural faithfulness,
and set $\tau_{\text{sem}}=0.65$; $\tau_{\text{len}}=1.25$ following RePro.
For reformatting, we fine-tune
Qwen3-1.7B on 10k labels from
Gemini 3.1 Flash-Lite as the faithful judge. The val. accuracy of our judge is 95\%.
Reward coefficients are $\lambda_{\text{quality}} = 1$, $\lambda_{\text{faithful}} = 1$, and
$\lambda_{\text{influence}} = 3$.
Empirically, \textsc{\ours} converges in three iterations.
More details and prompts are provided in Appendix~\ref{sec:app-setup}
and~\ref{sec:app-prompts}.

\section{Evaluation results}

In this section, we present main results (\S\ref{sec:results-main}), conduct ablations
(\S\ref{sec:results-ablation}), and analyze
pointwise faithfulness (\S\ref{sec:results-faithfulness}), distribution preservation
(\S\ref{sec:results-distribution}), and
model-awareness (\S\ref{sec:results-awareness}). Additional results and compute details
are provided in the
Appendix~\ref{sec:app-additional-results}.

\subsection{Main results}
\label{sec:results-main}


Table~\ref{tab:main} compares all methods across scales and data regimes.
\textsc{\ours} consistently outperforms all baselines.
At 400M, it improves over Repeat by 9.6\%, 7.5\%, and 9.3\% under
$\alpha{=}5\%$, $10\%$, and $15\%$, respectively.
As $\alpha$ increases, the gap between \textsc{\ours} and Unique Data narrows;
at $\alpha{=}15\%$, \textsc{\ours} even surpasses Unique Data
(0.5175 vs.\ 0.5133).
The gains remain notable at the 1.1B scale, where \textsc{\ours} outperforms
Repeat by 7.9\% and exceeds Unique Data (0.5584 vs.\ 0.5561) as well.
The same pattern holds when scaling up to 2B, where \textsc{\ours} improves over
Repeat by 7.6\% and again surpasses Unique Data (0.5859 vs.\ 0.5778),
indicating that the benefits persist rather than diminish as model scale grows.
Compared with RePro, \textsc{\ours} is especially strong on world knowledge,
as reformatting exposes factual content in structurally diverse forms and the
influence reward prioritizes knowledge the model has not yet absorbed.
These gains further extend beyond our evaluation suite, to held-out validation corpora
as well as continuation and more challenging reasoning benchmarks
(Appendix~\ref{sec:app-generalization}).
By contrast, QBSR and MATES yield modest gains, showing that selective
repetition alone may not be particularly helpful in data-bound regimes.

Furthermore, \textsc{\ours} delivers substantially stronger data efficiency than all baselines.
At 400M, the effective token multiplier grows from $3.4\times$ at $\alpha{=}5\%$ to
$5.2\times$ at $\alpha{=}10\%$ and $8.6\times$ at $\alpha{=}15\%$, where the
effective token count even exceeds the actual training tokens.
At 1.1B and 2B, the gain remains significant, with \textsc{\ours} reaching
$3.7\times$ and $3.4\times$ the effective tokens of Repeat, respectively.
Across all settings, the effective tokens of \textsc{\ours} are consistently
around $3\times$ those of RePro.
These results show that \textsc{\ours} generates more effective
pretraining tokens from a limited organic corpus, enabling more efficient utilization of
available data.
These gains also persist after post-training (Appendix~\ref{sec:app-sft}), and the
additional cost of our generation pipeline is analyzed in
Appendix~\ref{sec:app-compute}.

\subsection{Ablation studies}
\label{sec:results-ablation}

\begin{table}[t]
  \centering
  \renewcommand{\arraystretch}{1.3}
  \setlength{\tabcolsep}{3pt}
  \resizebox{1.0\columnwidth}{!}{%
    \begin{tabular}{l|c|cccc|cc|ccc|cc}
      \toprule
      & \textbf{\#Effective} &
      \multicolumn{4}{c|}{\textit{Commonsense Reasoning}} &
      \multicolumn{2}{c|}{\textit{Language Understanding}} &
      \multicolumn{3}{c|}{\textit{World Knowledge}} & & \textbf{Rec.} \\
      \textbf{Method} & \textbf{Tokens} &
      \textbf{CSQA} & \textbf{OBQA} & \textbf{PIQA} & \textbf{SIQA} &
      \textbf{HellaSwag} & \textbf{WinoG} &
      \textbf{ARC-e} & \textbf{ARC-c} & \textbf{SciQ} &
      \textbf{Avg} & \textbf{Ratio} \\
      \midrule
      \textit{Unique}     & \textit{48.0B} & \textit{0.3686} & \textit{0.3360} &
      \textit{0.6790} & \textit{0.4253} & \textit{0.4303} &
      \textit{0.5399} & \textit{0.6579} & \textit{0.3211} & \textit{0.8510} &
      \textit{0.5121} & \textit{100\%} \\
      \hline
      \rowcolor{blue!10}
      \textsc{\ours}       &
      \textbf{34.1B}$_{~\textcolor{red}{\textbf{5.2}\times}}$ & \textbf{0.3882} & \underline{0.3420} &
      \textbf{0.6649} & \textbf{0.4284} & \textbf{0.3868} & 0.5201 & 0.6123 &
      \textbf{0.3278} & 0.8540 & \textbf{0.5027} & \textbf{79\%} \\
      ~~w/o Rephrasing               & 14.3B$_{~\textcolor{red}{2.2\times}}$ & \underline{0.3759} &
      0.3280 & 0.6551 &
      0.4197 & 0.3518 &
      \underline{0.5280} & 0.6123 & 0.2742 & 0.8270 & 0.4858 & 41\% \\
      ~~w/o Reformatting                & 11.8B$_{~\textcolor{red}{1.8\times}}$ & 0.3423 &
      0.3200 & 0.6523 &
      0.4191 & 0.3825 &
      0.5272 & 0.5930 & 0.2809 & 0.8070 & 0.4805 & 29\% \\
      ~~w/o Quality Reward           & 11.3B$_{~\textcolor{red}{1.7\times}}$ & 0.3202 &
      0.3220 & 0.6529 &
      0.4089 & 0.3761 &
      0.5036 & 0.5860 & 0.3110 & 0.8320 & 0.4792 & 26\% \\
      ~~w/o Influence Reward         & 17.8B$_{~\textcolor{red}{2.7\times}}$ & 0.3702 &
      0.3340 & 0.6474 & 0.4253 & 0.3650 &
      0.5091 & 0.6035 & 0.3043 & \textbf{0.8580} & 0.4908 & 52\% \\
      ~~w/o Faithfulness Reward      & 13.8B$_{~\textcolor{red}{2.1\times}}$ & 0.3202 &
      0.3280 & 0.6415 &
      0.4202 & 0.3595 &
      0.5272 & \underline{0.6158} & 0.2943 & \underline{0.8570} & 0.4849 & 39\% \\
      ~~w/o RL Training              & 13.8B$_{~\textcolor{red}{2.1\times}}$ & 0.3260 &
      \textbf{0.3440} & 0.6404 &
      0.4191 & 0.3537 &
      0.5233 & 0.6105 & \underline{0.3144} & 0.8320 & 0.4848 & 39\% \\
      ~~w/o Data Merge     & \underline{18.3B}$_{~\textcolor{red}{2.8\times}}$ & 0.3726 & 0.3300 &
      \underline{0.6605} & 0.4227 &
      \underline{0.3840} & 0.5107 &
      0.6035 & 0.3110 & 0.8390 & \underline{0.4927} & \underline{57\%} \\
      \hline
      Nemotron-CC-HQ Prompt & 15.1B$_{~\textcolor{red}{2.3\times}}$ & 0.3292 & 0.3320 &
      0.6211 & \underline{0.4268} & 0.3682 &
      0.5138 & \textbf{0.6306} & \underline{0.3144} & 0.8490 & 0.4872 & 44\% \\
      Repeat                         & 6.6B$_{~{1.0\times}}$ & 0.3358 &
      0.3140 & 0.6442 &
      0.4135 & 0.3519 &
      \textbf{0.5304} & 0.5579 & 0.2676 & 0.7920 & 0.4675 & 0\% \\
      \bottomrule
    \end{tabular}
  }
  \caption{Ablation study on the 400M/$\alpha{=}10\%$ setting (0.8B organic tokens).}
  \label{tab:ablation}
  \vspace{-0.3cm}
\end{table}

We perform ablation studies in the 400M setting. Removing reformatting causes significant
drops in both commonsense reasoning ($-$2.6\%)
and world knowledge ($-$4.6\%), as reformatted outputs may expose factual content in
structured forms that reinforce both reasoning and knowledge learning.
Removing rephrasing causes a notable drop in language understanding ($-$4.0\% in
HellaSwag), indicating that
contextual comprehension benefits primarily from the lexical and structural
diversity that rephrasings provide.

Without the quality reward, the commonsense reasoning drops the most ($-$3.7\%),
suggesting that it encourages coherent generation useful for commonsense tasks.
Removing the influence reward produces a smaller but uniform drop across all
categories, confirming its role in pushing generation toward
content the model has yet to learn.
We also apply the Nemotron-CC-HQ~\citep{su2025nemotroncc} prompt to OLMo2-1B-Instruct.
Despite its improvements on knowledge tasks, the overall performance remains below
\textsc{\ours} and the
lack of faithfulness
guarantees may introduce \textit{distillation effects} where outputs reflect the
generator's knowledge rather than the organic content (see
\S\ref{sec:results-faithfulness}).
``w/o RL Training'' prompts the base generator with the same two operations directly, 
which lowers the recovery ratio from 79\% to 39\%, showing that the two operations alone are insufficient and RL is
essential to make the generator produce high-quality and faithful data.
``w/o Data Merge'' discards previous synthetic data and
trains only on the latest generation, resulting in a 22\% recovery ratio drop,
suggesting that accumulating data across iterations is more beneficial.
We further ablate the two models adopted in our generation pipeline, the semantic
faithfulness reward and the generator initialization, in
Appendix~\ref{sec:app-pipeline-ablation}.

\subsection{Pointwise faithfulness}
\label{sec:results-faithfulness}

\begin{figure*}[t]
  \centering
  \begin{subfigure}[t]
    {0.2342\textwidth}
    \raisebox{3.4pt}{\includegraphics[width=1.0\linewidth]{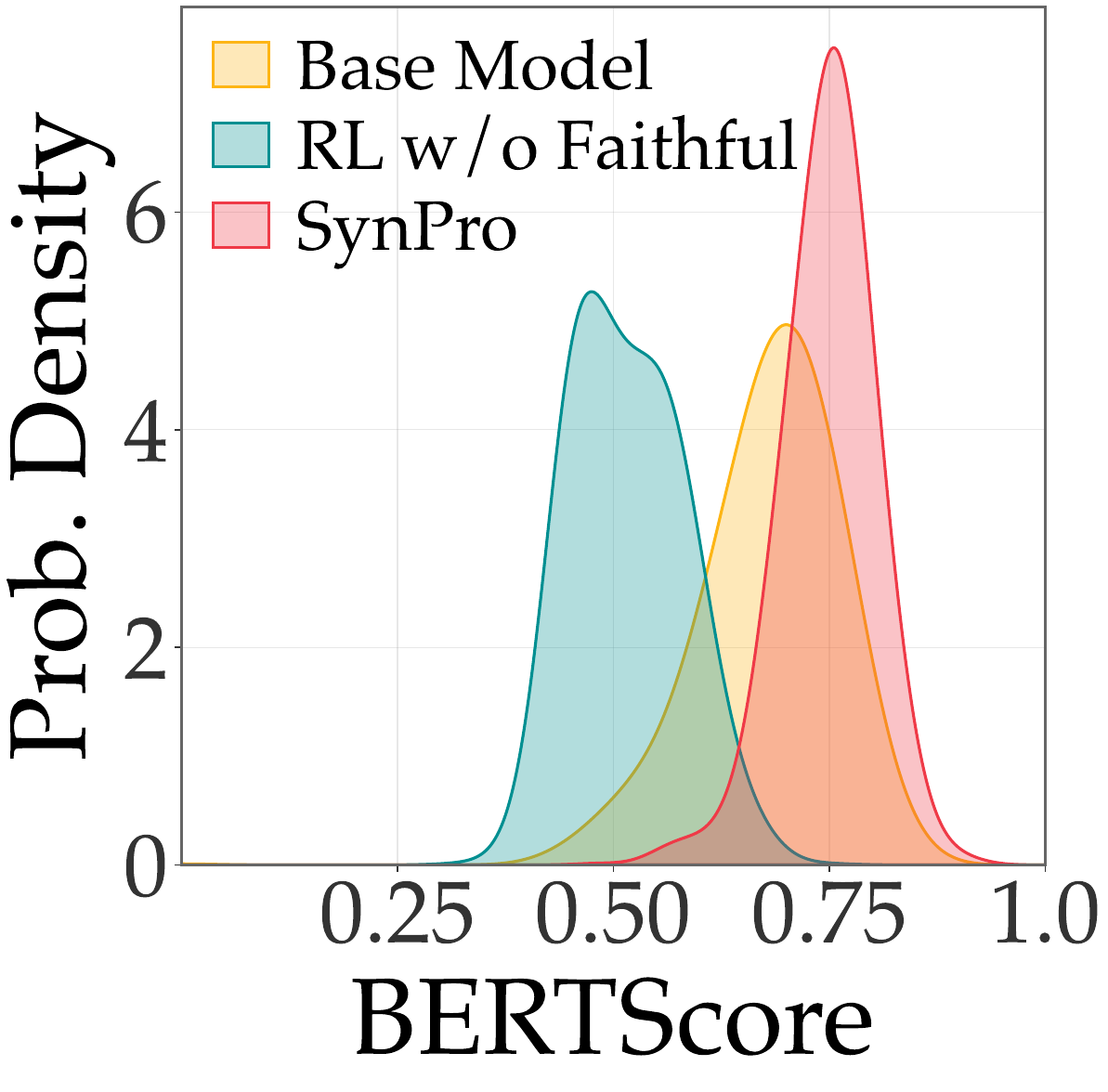}}
    \caption{BERTScore}
    \label{fig:faithfulness-a}
  \end{subfigure}
  ~
  \begin{subfigure}[t]
    {0.2422\textwidth}
    \centering
    \raisebox{4.4pt}{\includegraphics[width=1.0\linewidth]{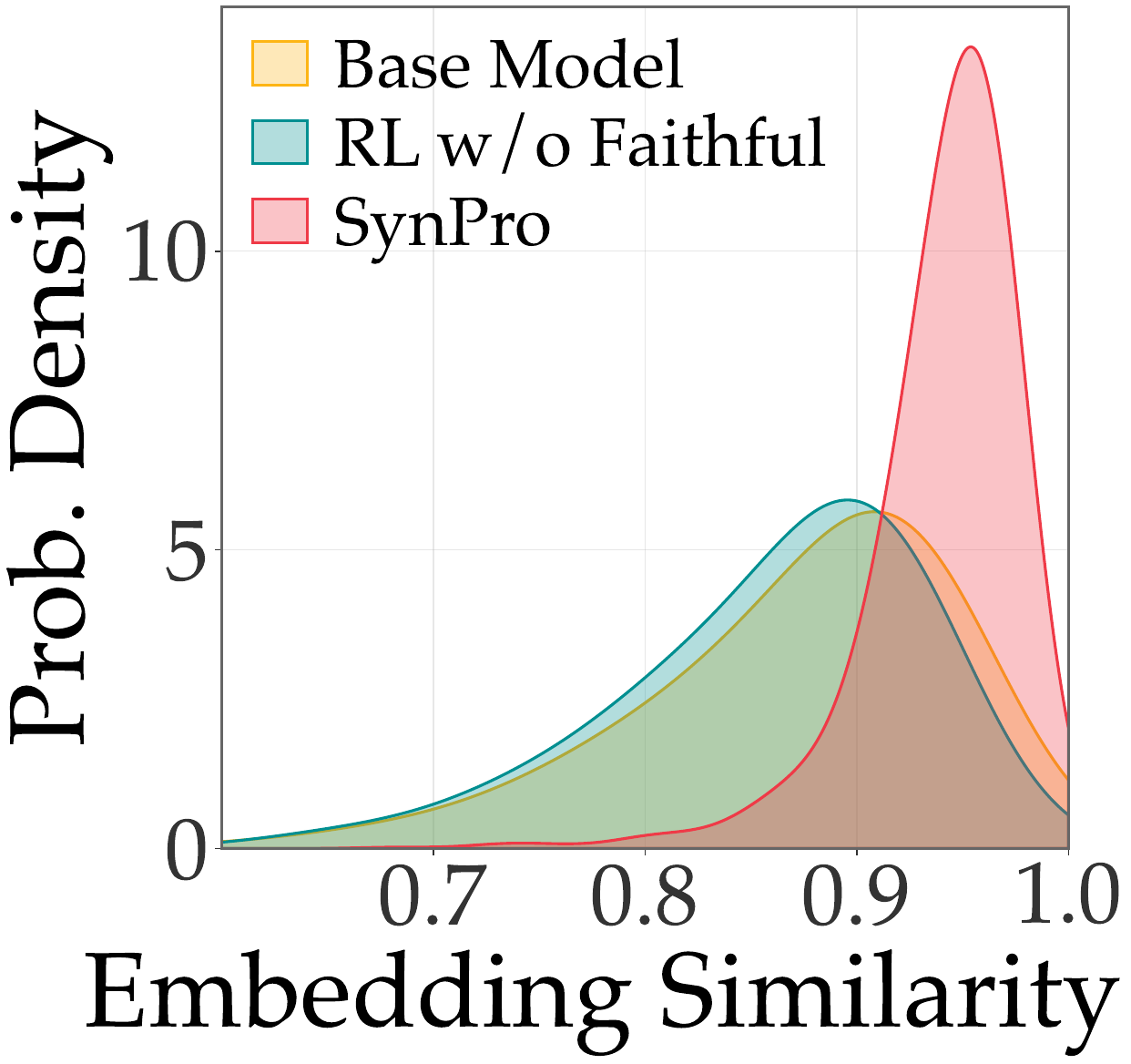}}
    \caption{jina-embeddings}
    \label{fig:faithfulness-b}
  \end{subfigure}
  ~
  \begin{subfigure}[t]
    {0.237\textwidth}
    \centering
    \raisebox{3.4pt}{\includegraphics[width=1.0\linewidth]{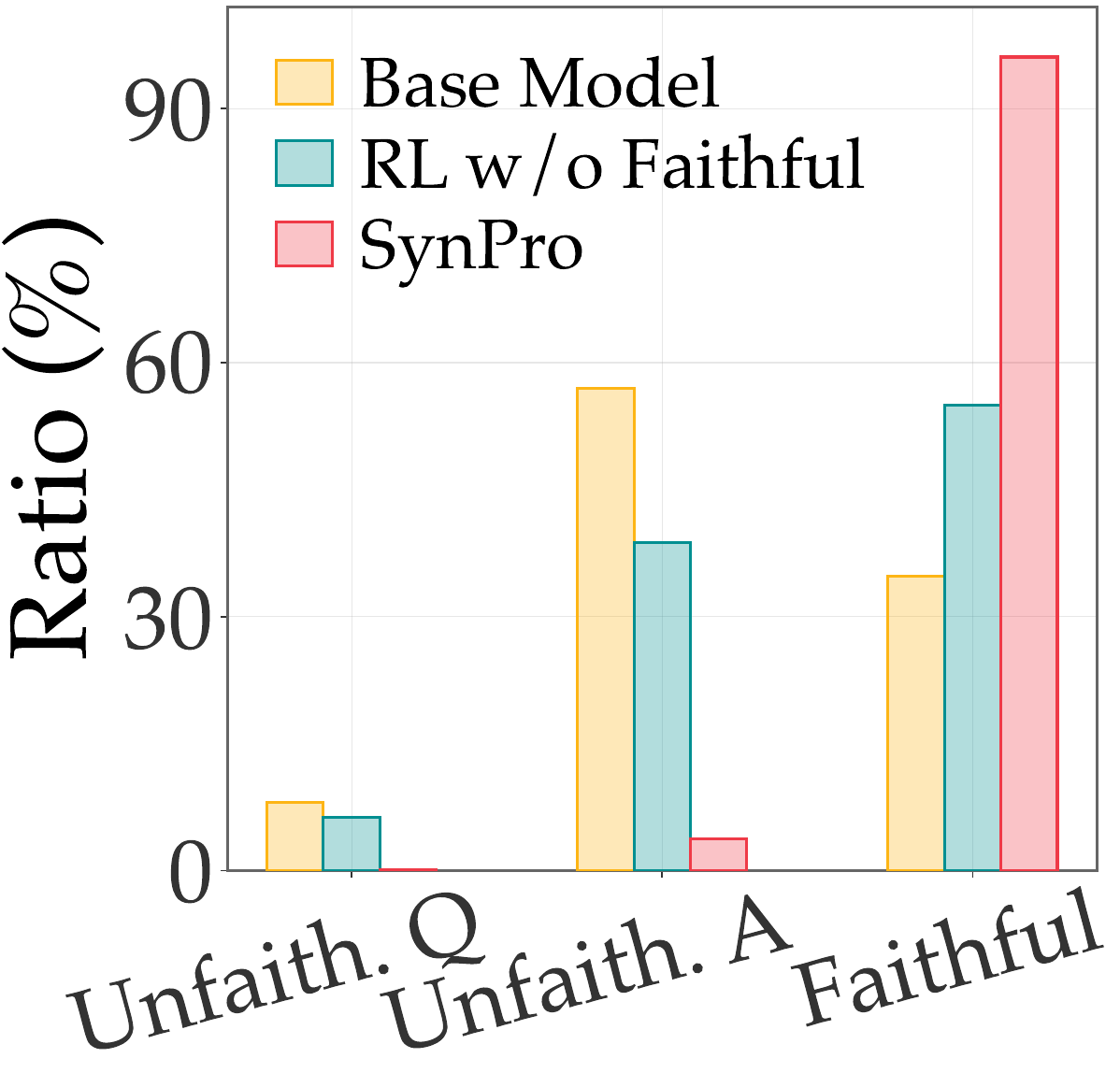}}
    \caption{Classification}
    \label{fig:faithfulness-c}
  \end{subfigure}
  ~
  \begin{subfigure}[t]
    {0.2342\textwidth}
    \centering
    \raisebox{2pt}{\includegraphics[width=1.0\linewidth]{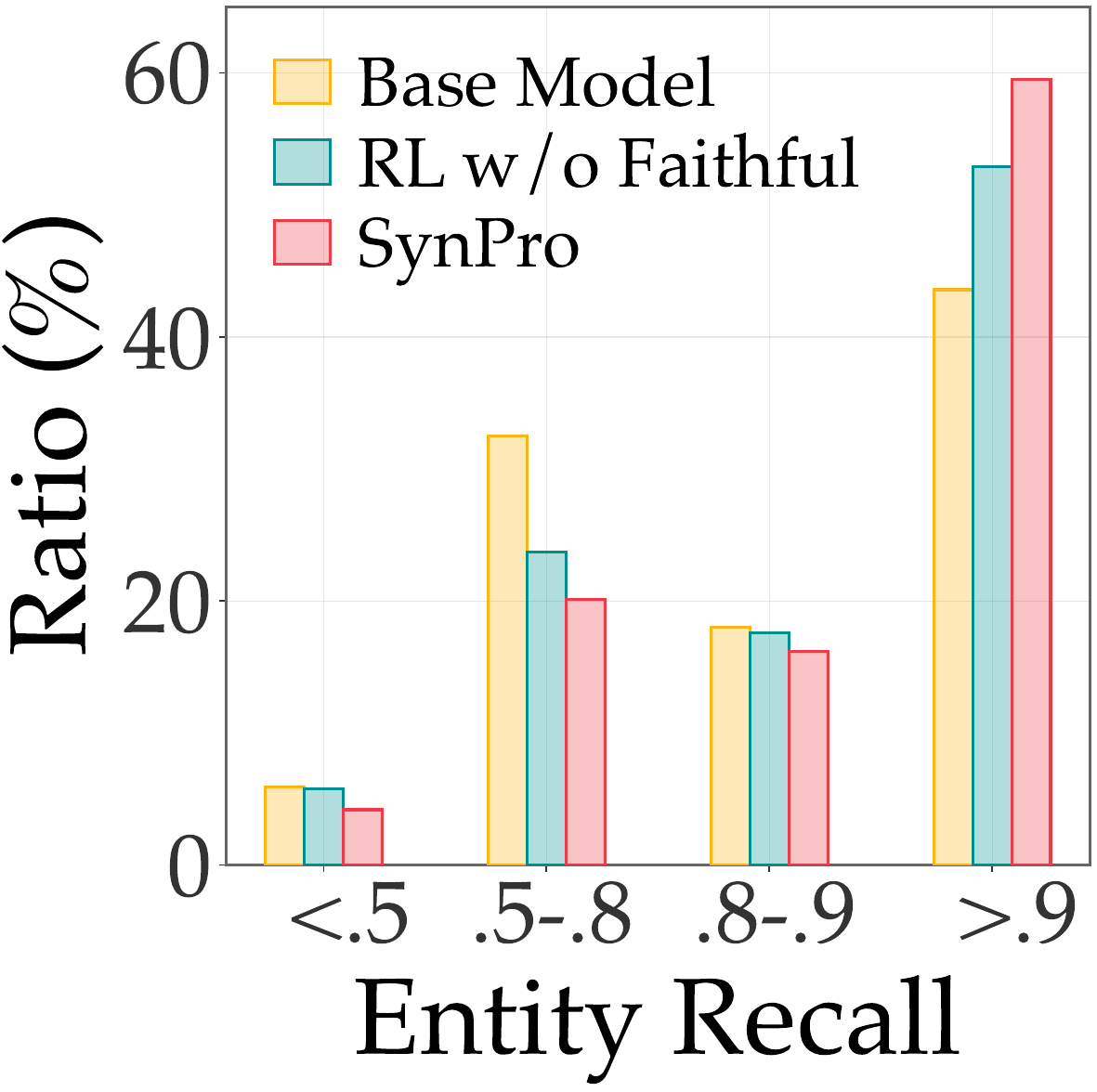}}
    \caption{Named Entity Recall}
    \label{fig:faithfulness-d}
  \end{subfigure}
  \caption{Faithfulness analysis on 1{,}000 randomly sampled organic documents not seen in RL.
    For rephrasing: (a)~BERTScore and (b) jina-embeddings similarity between original and
    rephrased text.
  For reformatting: (c)~faithfulness classification and (d)~named entity recall.}
  \label{fig:faithfulness}
  \vspace{-0.4cm}
\end{figure*}

\begin{figure*}[t]
  \centering
  \begin{subfigure}[t]
    {0.216\textwidth}
    \raisebox{7.5pt}{\includegraphics[width=1.0\linewidth]{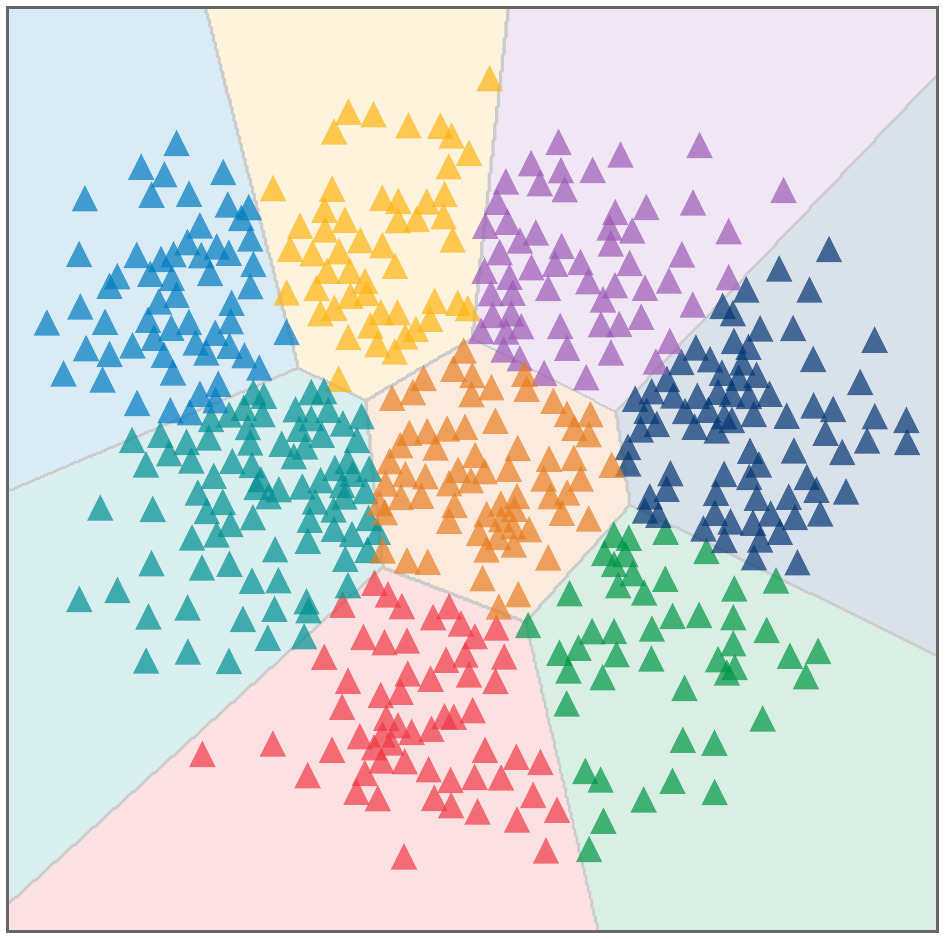}}
    \caption{Organic}
    \label{fig:distribution-a}
  \end{subfigure}
  ~
  \begin{subfigure}[t]
    {0.216\textwidth}
    \centering
    \raisebox{7.5pt}{\includegraphics[width=1.0\linewidth]{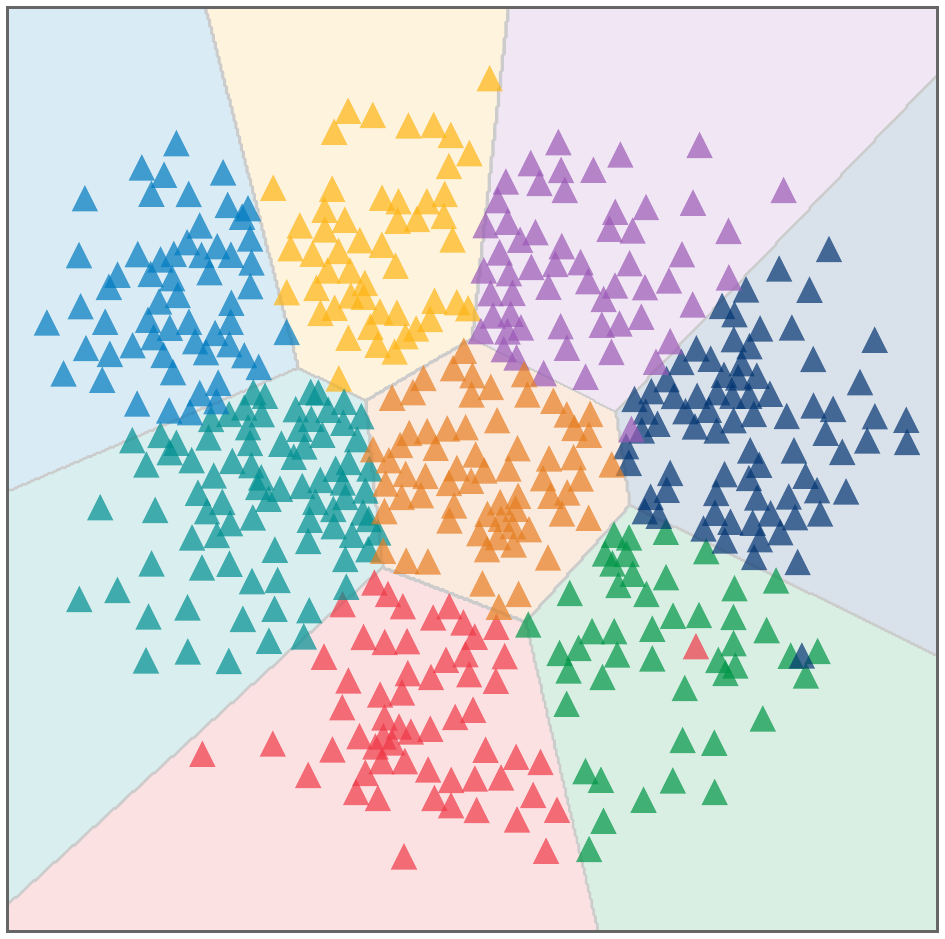}}
    \caption{\textsc{SynPro}}
    \label{fig:distribution-b}
  \end{subfigure}
  ~
  \begin{subfigure}[t]
    {0.216\textwidth}
    \centering
    \raisebox{7.5pt}{\includegraphics[width=1.0\linewidth]{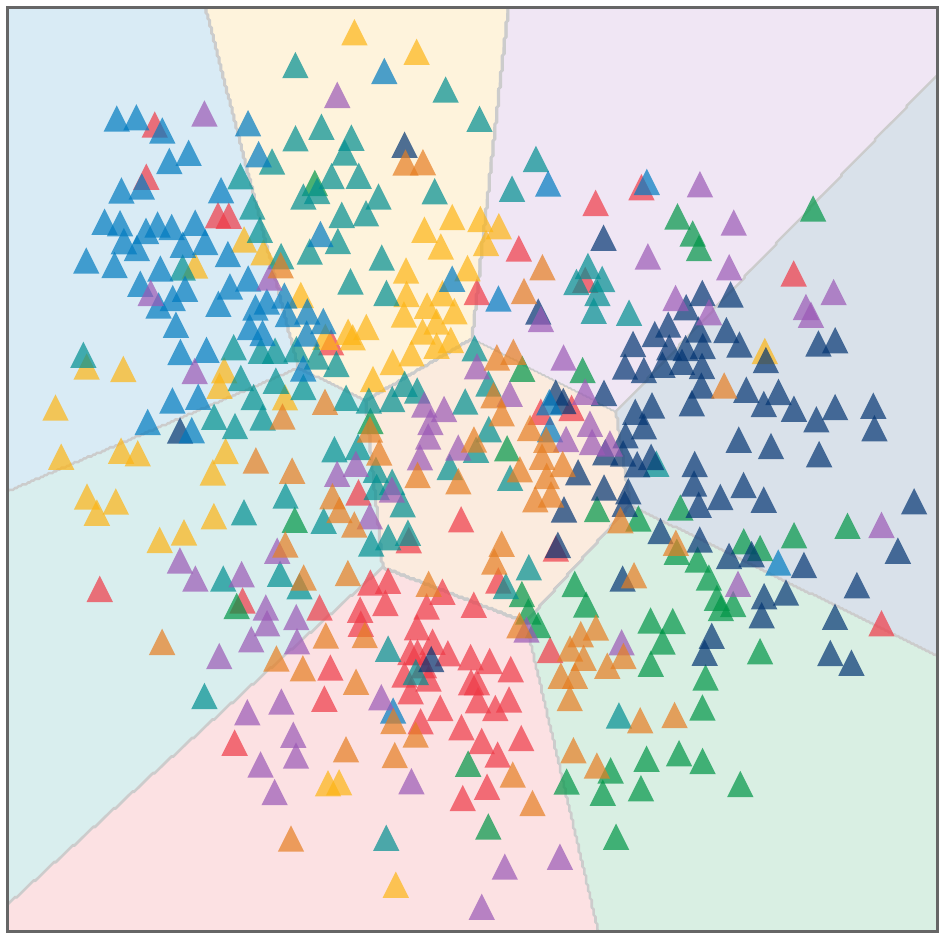}}
    \caption{Base Model}
    \label{fig:distribution-c}
  \end{subfigure}
  ~
  \begin{subfigure}[t]
    {0.248\textwidth}
    \centering
    \raisebox{0pt}{\includegraphics[width=1.0\linewidth]{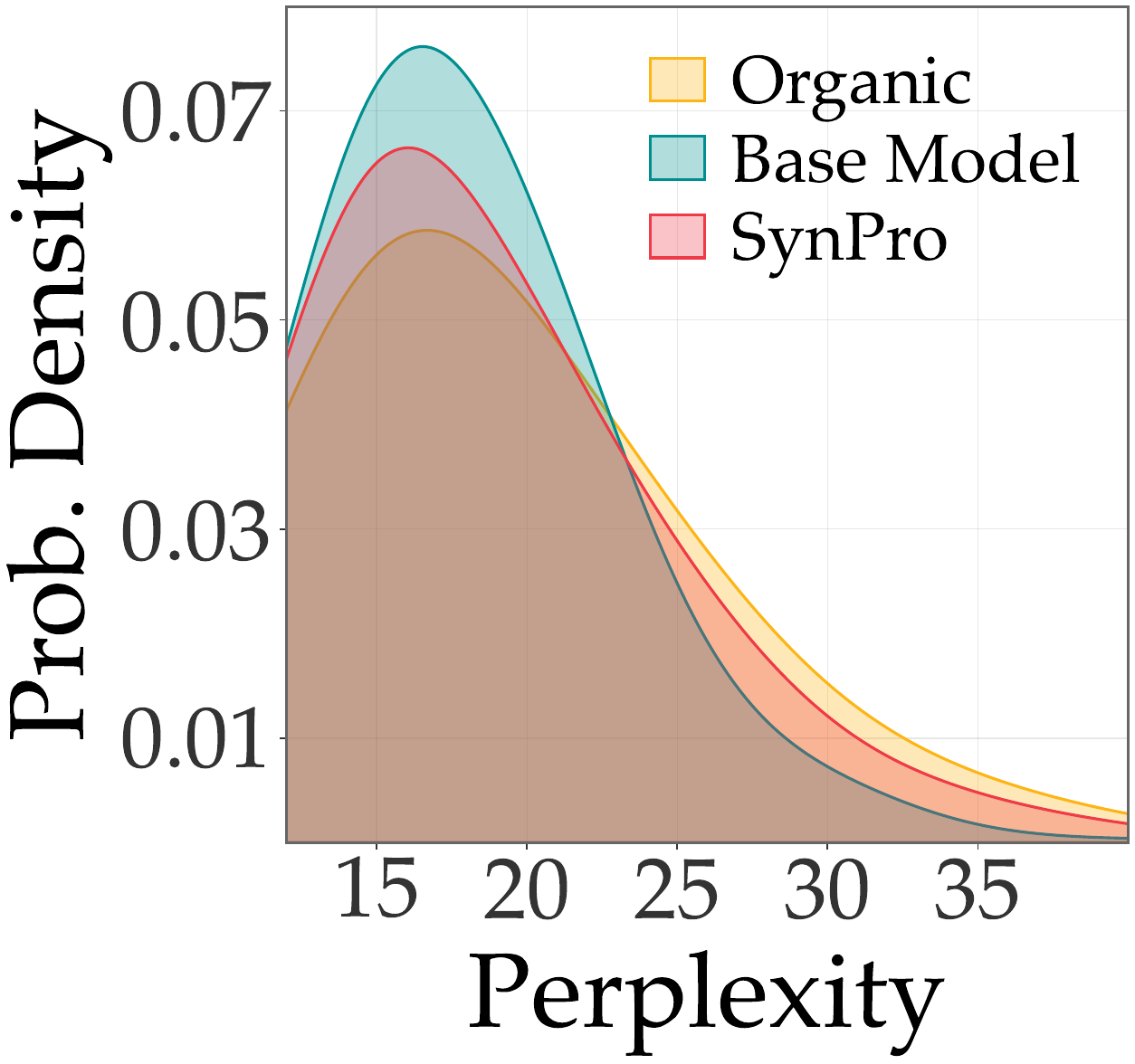}}
    \caption{Perplexity}
    \label{fig:distribution-d}
  \end{subfigure}
  \caption{Distribution preservation analysis. t-SNE illustration of Voronoi clusters,
    where each $\triangle$ denotes (a) one organic source, (b)
    \textsc{\ours} rephrasing, and (c) base-model rephrasing. (d)~Perplexity
  distributions from the 400M model trained on 61.2B unique tokens.}
  \label{fig:distribution}
  \vspace{-0.3cm}
\end{figure*}

In this analysis, we validate whether \textsc{\ours} preserves pointwise faithfulness to
the organic data, which is
critical for avoiding hallucinated content or distilled knowledge that
undermines generalization~\citep{yu2025repro}.
We randomly sample 1{,}000 organic documents not used in RL training and apply each
operation with three generators: the base model (OLMo2-1B-Instruct),
RL without the faithfulness reward, and \textsc{\ours}.

For rephrasing, we measure the semantic similarity between each organic document and the
rephrasing generated from it.
As shown in Figure~\ref{fig:faithfulness-a}, \textsc{\ours} achieves a mean BERTScore of
\textbf{0.75} with a tighter distribution, compared to 0.68 for the base model.
Notably, RL without the faithfulness reward drops to 0.52, confirming that optimizing
quality alone may harm faithfulness~\citep{yu2025repro}.
To verify generalization beyond BERTScore, we compute embedding similarity
using \textit{jina-embeddings-v5-text}~\citep{akram2026jina}
(Figure~\ref{fig:faithfulness-b}). Our generator achieves the highest
mean similarity (\textbf{0.94}) with the lowest variance among all baselines,
confirming genuine semantic preservation rather than reward overfitting.

For reformatting, our fine-tuned faithfulness judge (\S\ref{sec:setup}) categorizes each
output against its organic source as faithful (on-topic question with
a correct answer), unfaithful
(incorrect or unsupported) answer, or unfaithful (off-topic) question.
As shown in Figure~\ref{fig:faithfulness-c}, \textsc{\ours} achieves a \textbf{96.1\%}
faithfulness rate, compared to only 34.8\% for the base model and 55.0\% for RL without
the faithfulness reward.
The base model produces unfaithful answers 57.0\% of the time, indicating that without
targeted training, the generator frequently hallucinates content beyond the source document.
As an independent check, we extract named entities from the reformatted text with
\textit{BERT-base-NER}~\citep{devlin-etal-2019-bert} and compute recall against the
original. As shown in Figure~\ref{fig:faithfulness-d}, \textbf{76\%} of our samples exceed 80\%
recall versus 62\% for the base model, confirming that faithfulness generalizes beyond
the reward metric.

In summary, the faithfulness reward improves reward-aligned metrics
and generalizes to independent evaluations, confirming that \textsc{\ours}
produces faithful synthetic data.

\subsection{Distribution preservation}
\label{sec:results-distribution}


Beyond pointwise faithfulness, we examine whether the overall distribution of
synthetic data preserves the characteristics of the organic corpus, which is critical
for preventing model collapse and ensuring meaningful performance gains~\citep{shumailov2024nature}.

We first examine the preservation of the semantic distribution.
We embed 500 organic and rephrased texts with
\textit{jina-embeddings-v5-text},
apply k-means ($k{=}8$), and visualize the resulting
Voronoi regions on the t-SNE projections.
\textsc{\ours} (Figure~\ref{fig:distribution-b}) closely resembles the organic
distribution (Figure~\ref{fig:distribution-a}) and retains 99.2\% of cluster assignments,
confirming that our rephrasings preserve the semantics well.
By contrast, the base model (Figure~\ref{fig:distribution-c}) retains only 42.2\%,
with triangles scattered across mismatched regions, showing that without the
faithfulness constraint, the generator shifts the semantic distribution substantially.

We then focus on diversity preservation. Following~\citet{shumailov2024nature}, we
compute the perplexity of organic and
synthetic text from the 400M oracle model trained on 61.2B unique tokens.
As shown in Figure~\ref{fig:distribution-d}, \textsc{\ours} stays much closer to the
organic perplexity distribution than the base model, which produces a tighter,
lower-perplexity shape that does not capture the long tail of organic data.
This indicates that \textsc{\ours} better preserves the diversity of the organic corpus
rather than collapsing toward more predictable text.
Nevertheless, a smaller gap to the organic distribution still remains, as model-generated text may
carry intrinsic characteristics that narrow diversity, and we leave further exploration of this phenomenon to future work.


\subsection{Model-awareness analysis}
\label{sec:results-awareness}

\begin{figure*}[t]
  \centering
  \begin{subfigure}[t]
    {0.24\textwidth}
    \centering
    \raisebox{5pt}{\includegraphics[width=1.0\linewidth]{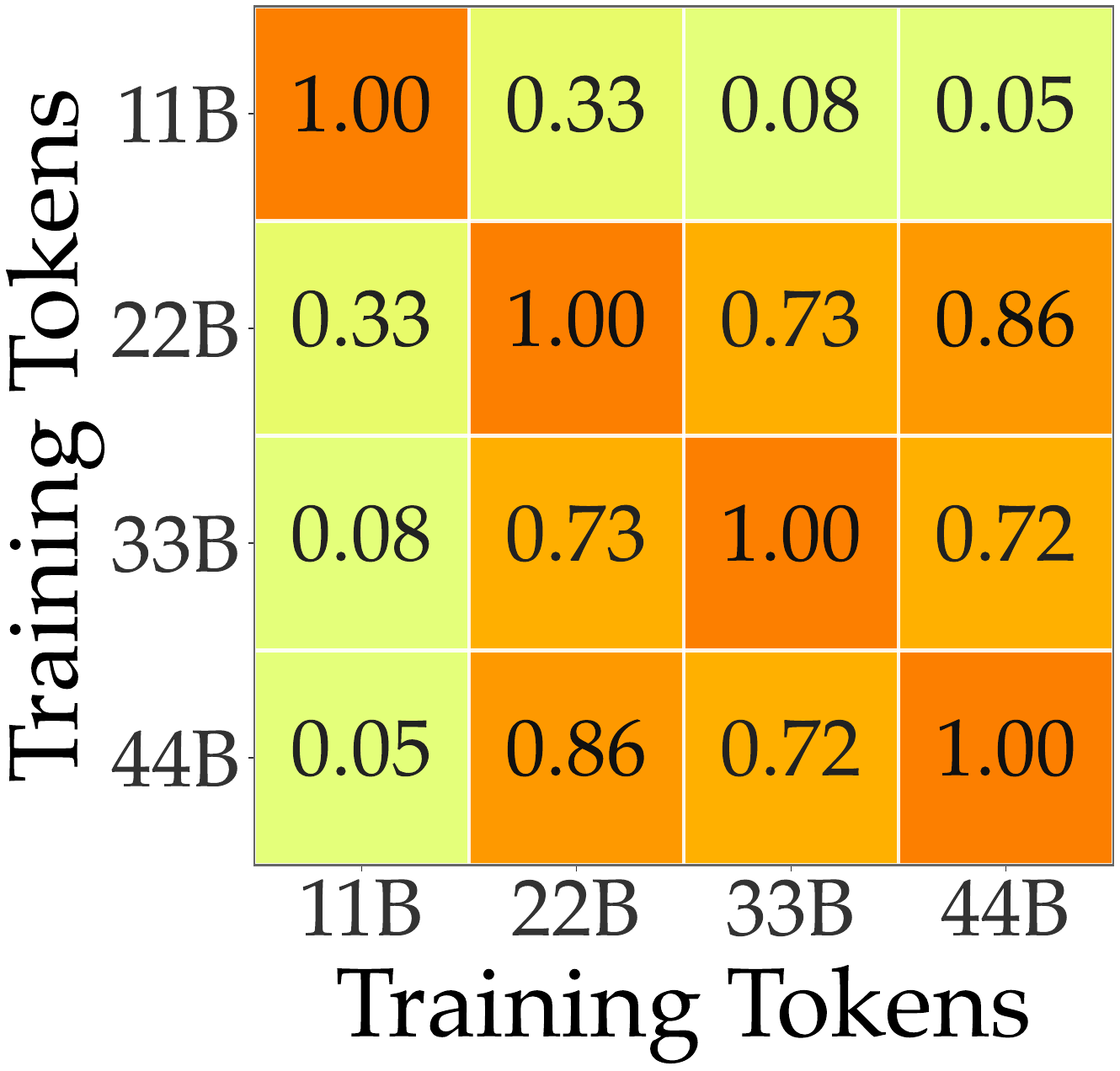}}
    \caption{Influence Correlation}
    \label{fig:awareness-a}
  \end{subfigure}
  ~
  \begin{subfigure}[t]
    {0.234\textwidth}
    \raisebox{5pt}{\includegraphics[width=1.0\linewidth]{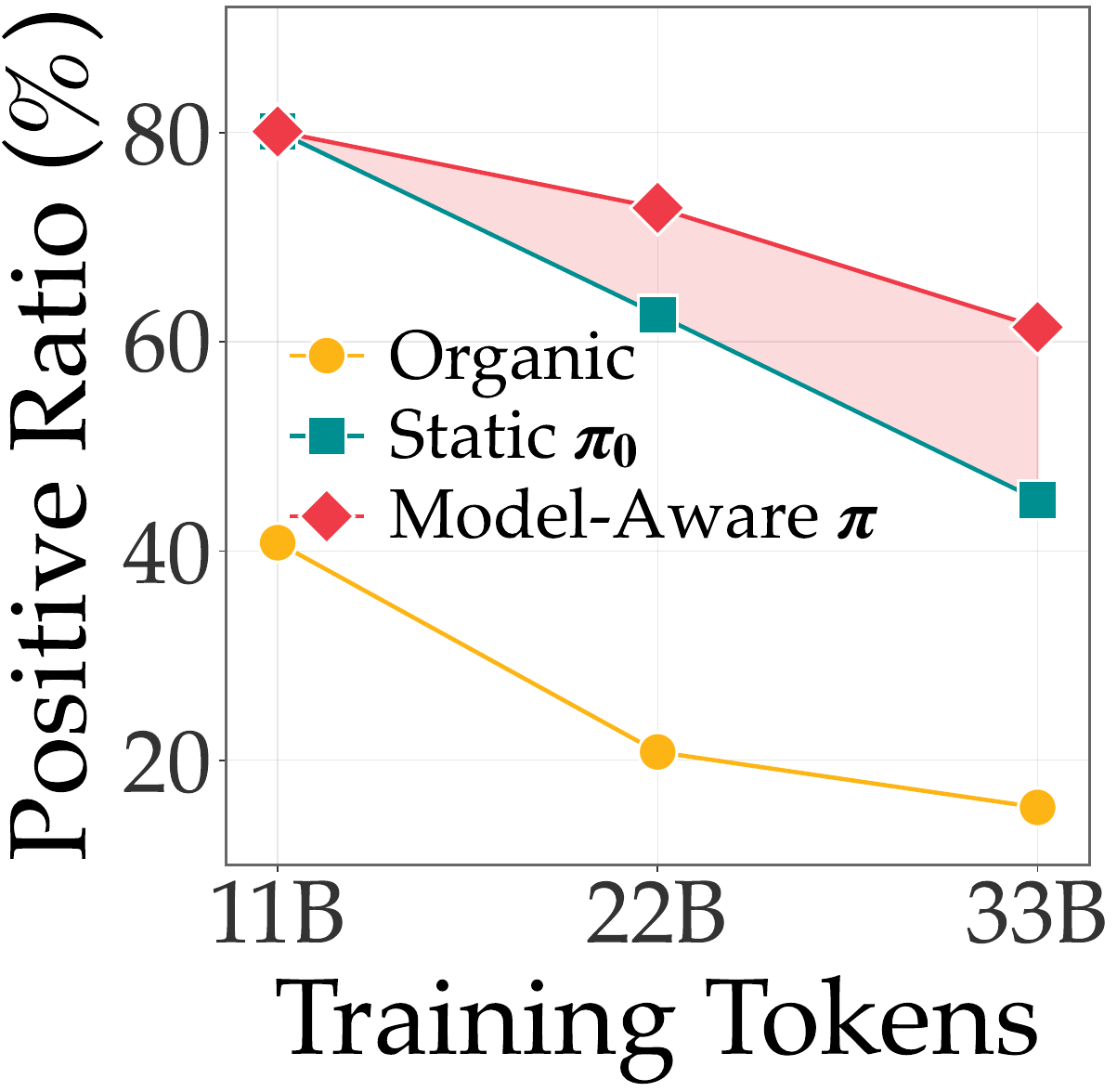}}
    \caption{Influence: Rephrase}
    \label{fig:awareness-b}
  \end{subfigure}
  ~
  \begin{subfigure}[t]
    {0.234\textwidth}
    \centering
    \raisebox{5pt}{\includegraphics[width=1.0\linewidth]{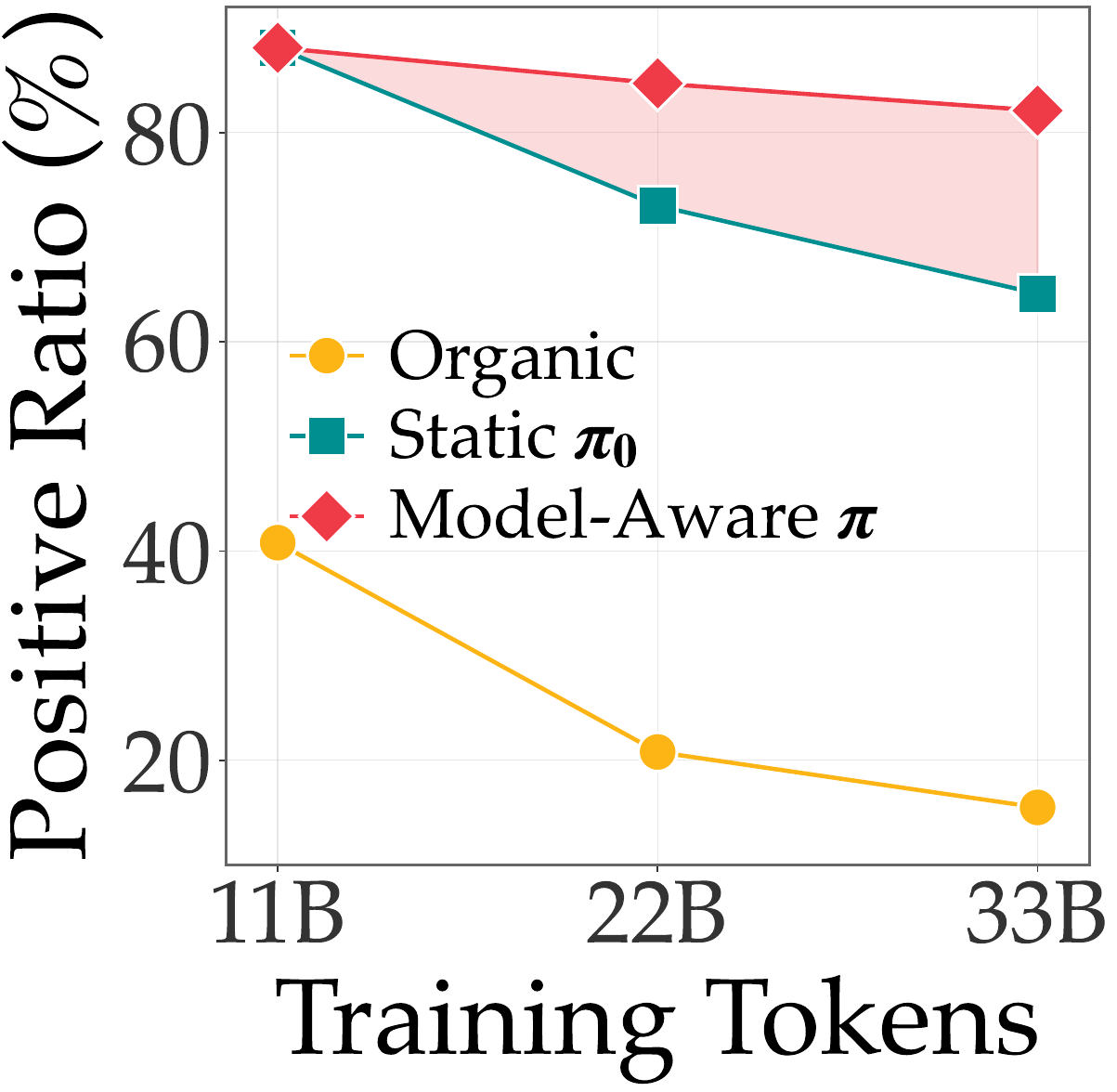}}
    \caption{Influence: Reformat}
    \label{fig:awareness-c}
  \end{subfigure}
  ~
  \begin{subfigure}[t]
    {0.237\textwidth}
    \centering
    \raisebox{0pt}{\includegraphics[width=1.0\linewidth]{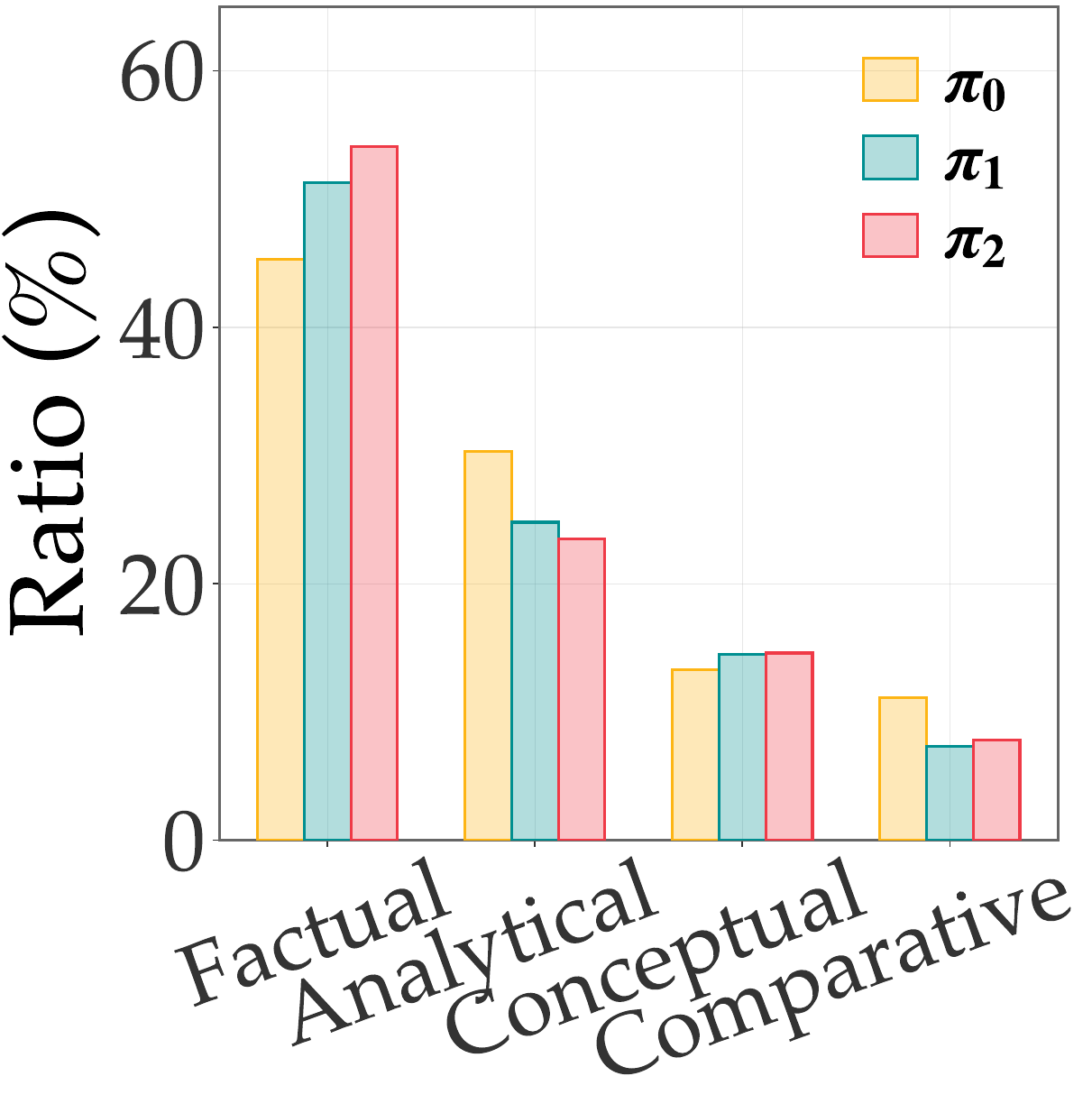}}
    \caption{Reformat Type}
    \label{fig:awareness-d}
  \end{subfigure}
  \caption{Model-awareness analysis on the 1.1B model. (a) Influence correlation and (b, c) positive
  influence ratio over pretraining. (d)~Reformat type distribution across iterations.}
  \label{fig:awareness}
  \vspace{-0.6cm}
\end{figure*}

Finally, we analyze how the influence reward shapes the generator's output across iterations.
First, we compute influence on the initial training data
$\mathcal{D}_{\text{org}} \cup \mathcal{D}_{\text{syn}}^{0}$. As shown in
Figure~\ref{fig:awareness-a}, the model's preferences shift substantially early in
pretraining, as reflected by the low influence correlation between 11B and later
checkpoints. 44B has a high influence correlation with 22B, reflecting that our model has
largely plateaued.
We further track the positive influence ratio (fraction of samples with positive data
influence, Eq.~\ref{eq:influence}) for both the static ($\pi_0$) and model-aware
($\pi_1$, $\pi_2$) policies.
Figures~\ref{fig:awareness-b} and~\ref{fig:awareness-c} show that both organic
and static synthetic data's
positive influence ratios drop rapidly as the model memorizes the repeated corpus.
By contrast, our model-aware policy maintains a consistently higher ratio after each
update, as the refreshed generator targets content the model has yet to learn, explaining
the sustained performance gains of \textsc{\ours}
over static approaches.

To examine how the generator adapts its outputs across iterations, we classify
reformatting
types via Gemini 3.1 Flash-Lite:
\textit{factual},
\textit{analytical},
\textit{conceptual}, and
\textit{comparative}.
Figure~\ref{fig:awareness-d} shows that from $\pi_0$ to $\pi_2$, the factual proportion
increases from 45.3\%
to 54.1\%, while analytical and comparative outputs decrease.
This suggests the influence reward steers the generator toward the model's factual gaps.
Appendix~\ref{sec:app-cases} shows cases to support this claim.





\section{Conclusion}


In this paper, we introduce \textsc{\ours}, an effective synthetic data generation
framework for the data-bound scaling regime.
Our results highlight two key insights.
First, organic data is underutilized, not exhausted. \textsc{\ours} helps the pretraining
model more thoroughly learn from a limited organic
corpus, matching performance achieved with much more unique data.
Second, faithfulness is essential for synthetic pretraining data, as grounding
outputs in the organic source enriches rather than distorts the training
distribution, while unconstrained generation risks collapse and distillation.
We hope \textsc{\ours} motivates future work to break the data wall and sustain LLM
scaling in the data-bound regime.



\section*{Acknowledgments}
We thank Amazon for funding Zichun Yu through the Amazon AI Ph.D. Fellowship Program.
We thank CMU Foundation and Language Model (FLAME) Center
for providing support of computational resources.

\section*{Ethics statement}
We use publicly available data and models, and we do not foresee significant ethical
concerns specific to this work beyond those already associated with language model
pretraining and synthetic data generation. However, as with other forms
of model-generated content, synthetic data may reflect biases introduced by the
generator itself. In our approach, we explicitly incorporate faithfulness
objectives into the reward design to mitigate such effects and encourage generated
data to closely preserve the distribution of the original data. While this does not
fully eliminate all risks associated with model-generated text, it provides a
principled mechanism for reducing unintended distortions during synthetic data
generation and helps maintain consistency between synthetic and original data
distributions.

\bibliography{bibliography}
\bibliographystyle{colm2026_conference}

\appendix
\clearpage
\startcontents[app]
\printcontents[app]{l}{1}{
\section*{Appendix Table of Contents}}

\section{Disclosure of LLM usage}

We truthfully disclose the following use of LLMs in this work.
First, LLMs are used as reward models in our method, including quality- and
faithfulness-related scoring components.
Second, LLM assistance was used for code implementation and figure scripting.
Third, LLM assistance was used to support literature search and review, as well
as to help draft parts of the paper text; however, references were added
manually by the authors.
All experiments were run manually by the authors, and all analyses were
performed and verified manually.
LLMs were not used to originate the core research ideas.

\section{\textsc{\ours} algorithm}

\textsc{\ours} algorithm is summarized in Algorithm~\ref{alg:synpro}.

\begin{algorithm}[ht]
  \caption{\textsc{\ours}: Model-Aware Synthetic Data Generation}
  \label{alg:synpro}
  \begin{algorithmic}[1]
    \Require Organic corpus $\mathcal{D}_{\text{org}}$, reference set
    $\mathcal{D}_{\text{ref}}$, initial generation policy $\pi_0$
    \Statex \textit{// Initialization}
    \State $\mathcal{D}_{\text{syn}}^{0} \gets \{\pi_0(p, x) \mid x \in
    \mathcal{D}_{\text{org}}\}$
    \State $\mathcal{D}_{\text{train}} \gets \mathcal{D}_{\text{org}} \cup
    \mathcal{D}_{\text{syn}}^{0}$
    \State $\mathcal{M}_0^{*} \gets$ random initialization
    \\
    \For{$i = 1, 2, \ldots$}
    \Statex \textit{~~~~~~// Stage 1: LM Pretraining}
    \State $\mathcal{M}_{i} \gets \mathcal{M}_{i-1}^{*}$, $\mathcal{L}_{0} \gets
    \mathcal{L}(\mathcal{D}_{\text{ref}} \mid \mathcal{M}_{i-1}^{*})$
    \For{$t = 1, 2, \ldots$}
    \State $\mathcal{M}_{i}^{'} \gets \mathcal{A}(\mathcal{M}_{i}, \mathcal{D}_{\text{train}})$
    \State $\mathcal{L}_{t} \gets \mathcal{L}(\mathcal{D}_{\text{ref}} \mid \mathcal{M}_{i}^{'})$
    \If{$t > 1$ and $\mathcal{L}_{t} \geq \min(\mathcal{L}_{t-1}, \mathcal{L}_{t-2})$}
    \State \textbf{break}
    \EndIf
    \State $\mathcal{M}_{i} \gets \mathcal{M}_{i}^{'}$
    \EndFor
    \State $\mathcal{M}_{i}^{*} \gets \mathcal{M}_{i}$
    \If{$\mathcal{L}(\mathcal{D}_{\text{ref}} \mid
      \mathcal{M}_{i}^{*}) \geq \mathcal{L}(\mathcal{D}_{\text{ref}} \mid
    \mathcal{M}_{i-1}^{*})$}
    \State \textbf{break}
    \EndIf
    \\
    \Statex \textit{~~~~~~// Stage 2: Policy update}
    \State $r_i(x, \tilde{x}) \gets \lambda_{\text{quality}}\,r_{\text{quality}}(\tilde{x}) +
    \lambda_{\text{faithful}}\,r_{\text{faithful}}(x, \tilde{x}) +
    \lambda_{\text{influence}}\,r_{\text{influence}}(\tilde{x} \mid \mathcal{M}_i^{*})$
    \State $\pi_{i} \gets \arg\max_{\pi}\;
    \mathbb{E}_{x \sim \mathcal{D}_{\text{org}},\,\tilde{x} \sim \pi(\cdot \mid p, x)}
    \!\left[r_i(x, \tilde{x})\right]$
    \\
    \Statex \textit{~~~~~~// Stage 3: Generation of new synthetic data}
    \State $\mathcal{D}_{\text{syn}}^{i} \gets \{\pi_{i}(p, x) \mid x \in
    \mathcal{D}_{\text{org}}\}$
    \State $\mathcal{D}_{\text{train}} \gets \mathcal{D}_{\text{train}} \cup
    \mathcal{D}_{\text{syn}}^{i}$
    \EndFor
  \end{algorithmic}
\end{algorithm}

\section{Derivation of the influence approximation}
\label{sec:app-influence}

Following MATES~\citep{yu2024mates}, we first formulate the oracle data influence of a
sample $x$ by the change in reference loss after training on $x$:
\begin{align}
  \mathcal{I}(x \mid \mathcal{M}_i^{*})
  =
  \mathcal{L}(\mathcal{D}_{\text{ref}} \mid \mathcal{A}(\mathcal{M}_i^{*}, x))
  -
  \mathcal{L}(\mathcal{D}_{\text{ref}} \mid \mathcal{M}_i^{*}).
  \label{eq:app-influence-oracle}
\end{align}
Define
\begin{align}
  g_x
  &= \nabla_{\mathcal{M}}\mathcal{L}(x \mid \mathcal{M}_i^{*}), \\
  g_{\text{ref}}
  &= \nabla_{\mathcal{M}}\mathcal{L}(\mathcal{D}_{\text{ref}} \mid \mathcal{M}_i^{*}).
\end{align}
Approximating one update on $x$ and one update on $\mathcal{D}_{\text{ref}}$ by
gradient steps with step size $\eta$,
\begin{align}
  \mathcal{A}(\mathcal{M}_i^{*}, x)
  &\approx \mathcal{M}_i^{*} - \eta g_x, \\
  \mathcal{A}(\mathcal{M}_i^{*}, \mathcal{D}_{\text{ref}})
  &\approx \mathcal{M}_i^{*} - \eta g_{\text{ref}}.
\end{align}
and using first-order Taylor expansion around $\mathcal{M}_i^{*}$,
\begin{align}
  \mathcal{I}(x \mid \mathcal{M}_i^{*})
  &=
  \mathcal{L}(\mathcal{D}_{\text{ref}} \mid \mathcal{A}(\mathcal{M}_i^{*}, x))
  -
  \mathcal{L}(\mathcal{D}_{\text{ref}} \mid \mathcal{M}_i^{*})
  \notag \\
  &\approx
  \nabla_{\mathcal{M}}\mathcal{L}(\mathcal{D}_{\text{ref}} \mid \mathcal{M}_i^{*})^{\top}
  \left(\mathcal{A}(\mathcal{M}_i^{*}, x) - \mathcal{M}_i^{*}\right) \notag \\
  &\approx -\eta\, g_x^{\top} g_{\text{ref}} \notag \\
  &\approx
  -\nabla_{\mathcal{M}}\mathcal{L}(x \mid \mathcal{M}_i^{*})^{\top}
  \left(\mathcal{M}_i^{*} - \mathcal{A}(\mathcal{M}_i^{*}, \mathcal{D}_{\text{ref}})\right)
  \notag \\
  &\approx \mathcal{L}(x \mid \mathcal{A}(\mathcal{M}_i^{*}, \mathcal{D}_{\text{ref}})) -
  \mathcal{L}(x \mid \mathcal{M}_i^{*}).
  \label{eq:app-influence-oracle-approx}
\end{align}
which matches the mirrored influence view of Forward-INF~\citep{ko2024mirrored}. The
practical benefit is that we only need one update on $\mathcal{D}_{\text{ref}}$ to
form $\mathcal{A}(\mathcal{M}_i^{*}, \mathcal{D}_{\text{ref}})$, after which scoring
each candidate sample only requires evaluating
$\mathcal{L}(x \mid \mathcal{M}_i^{*})$ and
$\mathcal{L}(x \mid \mathcal{A}(\mathcal{M}_i^{*}, \mathcal{D}_{\text{ref}}))$,
i.e., forward inference only.
where we take the negative oracle as the influence reward used in
\S\ref{sec:method-operations} since higher reward should denote better data:
\begin{align}
  r_{\text{influence}}(x \mid \mathcal{M}_i^{*})
  =
  \mathcal{L}(x \mid \mathcal{M}_i^{*})
  -
  \mathcal{L}(x \mid \mathcal{A}(\mathcal{M}_i^{*}, \mathcal{D}_{\text{ref}}))
  \approx -\mathcal{I}(x \mid \mathcal{M}_i^{*}).
\end{align}

\section{Experimental details}
\label{sec:app-setup}

We provide training details in Table~\ref{tab:exp-details}
and implementation details of \textsc{\ours} in Table~\ref{tab:impl-details}.

\begin{table}[ht]
  \centering
  \renewcommand{\arraystretch}{1.3}
  \setlength{\tabcolsep}{3pt}
  \resizebox{0.7\linewidth}{!}{%
    \begin{tabular}{l|ccc}
      \toprule \textbf{Hyperparameter} & \textbf{400M LM} & \textbf{1.1B LM} &
      \textbf{Synthetic Data Generator} \\
      \hline
      Steps                     & -                & -         & 600 \\
      Batch size                & 512              & 512       & 24 \\
      Sequence length           & 2048             & 2048      & 4096 (2048i + 2048o) \\
      Max learning rate         & 6e-4             & 4e-4      & 1e-6 \\
      Optimizer                 & AdamW            & AdamW     & AdamW \\
      Scheduler                 & WSD              & WSD       & Cosine \\
      \bottomrule
    \end{tabular}
  }
  \caption{Training details.}
  \label{tab:exp-details}
\end{table}

\begin{table}[t]
  \centering
  \renewcommand{\arraystretch}{1.3}
  \setlength{\tabcolsep}{3pt}
  \resizebox{0.95\linewidth}{!}{%
    \begin{tabular}{lr}
      \toprule
      \textbf{Component} & \textbf{Setting} \\
      \hline
      Reference set $\mathcal{D}_{\text{ref}}$ & FLAN~\citep{wei2022flan} \\
      Initial policy $\pi_0$ & OLMo2-1B-Instruct \\
      Quality reward model & DataMan~\citep{peng2025dataman} \\
      Rephrasing faithfulness model & Qwen3-1.7B~\citep{yang2025qwen3} \\
      Rephrasing faithfulness method & Few-shot prompting for structural faithfulness \\
      Reformat faithfulness model & Qwen3-1.7B \\
      Reformat faithfulness method & Fine-tuned on 10{,}000 distilled from Gemini 3.1 Flash-Lite \\
      $\tau_{\text{sem}}$ & 0.65 \\
      $\tau_{\text{len}}$ & 1.25 \\
      $\lambda_{\text{quality}}$ & 1 \\
      $\lambda_{\text{faithful}}$ & 1 \\
      $\lambda_{\text{influence}}$ & 3 \\
      Iterations to converge & 3 \\
      \hline
      Generator training algorithm & GRPO~\citep{shao2024deepseekmath} \\
      GRPO clipping $\epsilon$ & 0.2 \\
      GRPO $\beta$ & 0.005 \\
      GRPO rollouts per input & 8 \\
      \hline
      Generator inference framework & vllm~\citep{kwon2023efficient} \\
      Decoding temperature & 1.0 \\
      Decoding top-$p$ & 0.9 \\
      \bottomrule
    \end{tabular}
  }
  \caption{Implementation details of \textsc{\ours}.}
  \label{tab:impl-details}
\end{table}

\section{Additional results}
\label{sec:app-additional-results}
In this section, we report generalization results
(\S\ref{sec:app-generalization}), verify compatibility with post-training
(\S\ref{sec:app-sft}), discuss compute cost (\S\ref{sec:app-compute}), ablate the models
used in our generation pipeline (\S\ref{sec:app-pipeline-ablation}), analyze the
quality of our synthetic data (\S\ref{sec:app-quality}), and examine RL training
dynamics (\S\ref{sec:app-rl-dynamics}).

\begin{figure*}[b]
  \centering
  \raisebox{0pt}{\includegraphics[width=0.47\textwidth]{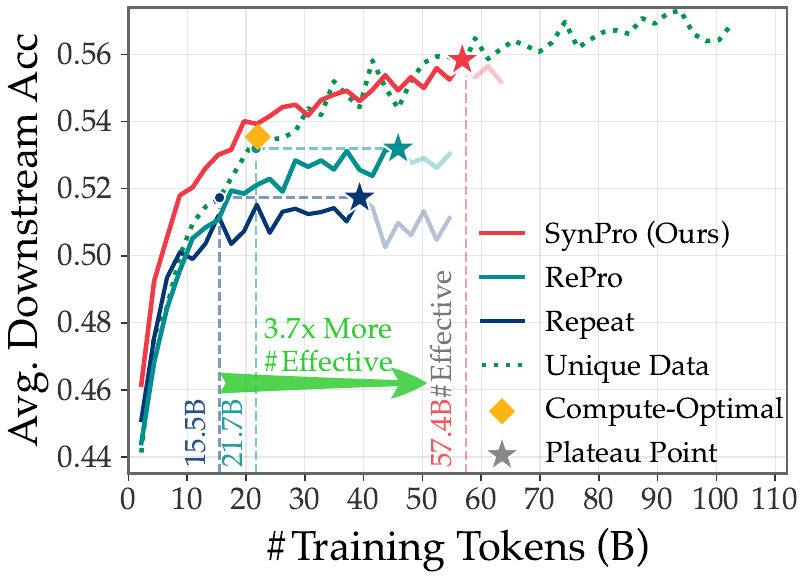}}
  \caption{1B model \& 2.2B unique organic tokens}
  \label{fig:1b-curves}
\end{figure*}

\subsection{Generalization results}
\label{sec:app-generalization}

\begin{table}[t]
  \centering
  \renewcommand{\arraystretch}{1.3}
  \setlength{\tabcolsep}{3pt}
  \resizebox{1.0\textwidth}{!}{%
    \begin{tabular}{l|ccc|cccc}
      \toprule
      & \multicolumn{3}{c|}{\textbf{Validation Corpora}} &
      \multicolumn{4}{c}{\textbf{Generalization Tasks}} \\
      \textbf{Method} &
      \textbf{DCLM-val (CE)$\downarrow$} & \textbf{C4-val (CE)$\downarrow$} &
      \textbf{Pile-val (CE)$\downarrow$} &
      \textbf{LAMBADA (CE)$\downarrow$} & \textbf{LAMBADA (Acc)$\uparrow$} &
      \textbf{COPA$\uparrow$} & \textbf{MMLU$\uparrow$} \\
      \hline
      Repeat & 2.855 & 3.359 & 2.923 & 0.6645 & 0.3577 & 0.6300 & 0.3049 \\
      RePro & 2.815 & 3.299 & 2.849 & 0.6136 & 0.3771 & 0.7200 & 0.3157 \\
      \rowcolor{blue!10}
      \textsc{\ours} & \textbf{2.771} & \textbf{3.242} & \textbf{2.772} &
      \textbf{0.5823} & \textbf{0.4007} & \textbf{0.7300} & \textbf{0.3326} \\
      \bottomrule
    \end{tabular}
  }
  \caption{Generalization results on the 1.1B/$\alpha{=}10\%$ setting.
    CE denotes cross-entropy loss (lower is better) and Acc denotes accuracy
    (higher is better). MMLU is evaluated with 5 shots.
    DCLM-val is in-distribution, while C4-val and Pile-val are
  out-of-distribution.}
  \label{tab:generalization}
\end{table}

To verify that the gains of \textsc{\ours} come from genuinely better learning, we
further evaluate the 1.1B models on held-out validation corpora and more downstream benchmarks (Table~\ref{tab:generalization}).

We first report validation loss on three widely used corpora,
DCLM-val (in-distribution), C4-val and Pile-val (out-of-distribution).
\textsc{\ours} achieves the lowest loss on all three corpora, reducing
DCLM-val from 2.855 to \textbf{2.771}, C4-val from 3.359 to \textbf{3.242}, and
Pile-val from 2.923 to \textbf{2.772} relative to Repeat. 
These results indicate that \textsc{\ours} helps the model 
better learn and generalize from limited organic data.

The gains of \textsc{\ours} also
generalize well to continuation tasks such as
LAMBADA~\citep{paperno2016lambada} and
COPA~\citep{roemmele2011choice}, even though these are
not directly targeted by our synthetic data operation.
\textsc{\ours} consistently improves over Repeat and RePro on both tasks, reducing
LAMBADA cross-entropy from 0.6645 to \textbf{0.5823} and improving COPA accuracy
from 63.0\% to \textbf{73.0\%} relative to Repeat.
These gains indicate that the benefits of \textsc{\ours} extend beyond the
evaluation tasks in Table~\ref{tab:main}, improving broader contextual prediction
and causal reasoning rather than merely fitting the specific formats used during
synthetic data generation.

We finally evaluate our 1.1B checkpoints on 5-shot MMLU~\citep{hendryckstest2021}, a
widely used and challenging knowledge-reasoning benchmark.
\textsc{\ours} improves over Repeat by \textbf{2.8\%} absolute, doubling the gain of
RePro, our strongest baseline.
These gains indicate that \textsc{\ours} further strengthens the model's capability on
more complex reasoning tasks.

\subsection{Compatibility with post-training}
\label{sec:app-sft}

\begin{table}[t]
  \centering
  \renewcommand{\arraystretch}{1.3}
  \setlength{\tabcolsep}{3pt}
  \resizebox{1.0\textwidth}{!}{%
    \begin{tabular}{l|cccc|cc|ccc|c}
      \toprule
      & \multicolumn{4}{c|}{\textit{Commonsense Reasoning}} &
      \multicolumn{2}{c|}{\textit{Language Understanding}} &
      \multicolumn{3}{c|}{\textit{World Knowledge}} & \\
      \textbf{Method} &
      \textbf{CSQA} & \textbf{OBQA} & \textbf{PIQA} & \textbf{SIQA} &
      \textbf{HellaSwag} & \textbf{WinoG} &
      \textbf{ARC-e} & \textbf{ARC-c} & \textbf{SciQ} & \textbf{Avg} \\
      \midrule
      Repeat & 0.369 & 0.342 & 0.693 & 0.430 & 0.459 & 0.530 & 0.653 & 0.334 &
      0.845 & 0.517 \\
      ~~+ SFT & 0.417 & 0.346 & 0.693 & 0.451 & 0.453 & 0.534 & 0.665 & 0.341 &
      0.864 & 0.529 \\
      \hline
      RePro & 0.403 & 0.346 & 0.694 & 0.437 & 0.489 & 0.545 & 0.665 & 0.331 &
      0.877 & 0.532 \\
      ~~+ SFT & 0.433 & 0.334 & \underline{0.699} & \underline{0.468} & 0.481 &
      \underline{0.559} &
      0.668 & 0.348 & 0.892 & 0.542 \\
      \hline
      \rowcolor{blue!10}
      \textsc{\ours} & \underline{0.459} & \textbf{0.352} & 0.691 & \underline{0.468} &
      \textbf{0.492} & 0.539 & \underline{0.702} & \textbf{0.415} & \underline{0.909} &
      \underline{0.558} \\
      \rowcolor{blue!10}
      ~~+ SFT & \textbf{0.493} & \underline{0.350} & \textbf{0.700} & \textbf{0.472} &
      \underline{0.491} & \textbf{0.563} & \textbf{0.714} & \underline{0.391} &
      \textbf{0.923} & \textbf{0.566} \\
      \bottomrule
    \end{tabular}
  }
  \caption{Downstream accuracy before and after supervised fine-tuning (SFT) on
  FLAN~\citep{wei2022flan}, on the 1.1B/$\alpha{=}10\%$ setting.}
  \label{tab:sft}
\end{table}

To examine whether the benefits of better pretraining data persist after post-training,
we fine-tune the 1.1B checkpoints of Repeat, RePro, and \textsc{\ours} on 
FLAN, a standard open-source SFT dataset.
As shown in Table~\ref{tab:sft}, \textsc{\ours} retains a \textbf{3.7\%} absolute gain
over Repeat after SFT, close to the \textbf{4.1\%} gain before SFT, and more than
doubles the gain of RePro (1.3\%).
These results indicate that the advantage of our synthetic data is not washed out by
post-training, and that a stronger pretrained backbone remains valuable after SFT.

\subsection{Compute cost}
\label{sec:app-compute}

Table~\ref{tab:compute} reports the H100 GPU hours for each method under the same number
of training tokens.
Compared with simple repetition, \textsc{\ours} incurs additional synthesis cost,
but the \emph{relative} overhead decreases as the pretraining model scales:
the total cost is $13.6/2.7 \approx 5.0\times$ Repeat at 400M, but only $49.0/19.2 \approx 2.6\times$ at 1.1B.
This trend is expected because pretraining becomes increasingly dominant at larger
model sizes, while our synthesis pipeline relies on a relatively lightweight 1B model,
so the extra cost of our method is progressively amortized.

Table~\ref{tab:cost-breakdown} further decomposes the total cost of \textsc{\ours} over
the whole pipeline into generator RL training, generator inference, and language model
pretraining.
As the model size scales up, pretraining increasingly dominates the total cost
(from 73.3\% at 400M to 80.8\% at 1.1B), so our entire data generation pipeline takes
only 26.7\% of the total cost at 400M and 19.2\% at 1.1B.
More importantly, our focus is data-bound pretraining, where naively spending
compute on repeated passes over the same small organic corpus quickly hits the data
wall and yields diminishing returns.
In this regime, paying moderate additional compute to generate grounded and
model-aware synthetic data is acceptable, since it unlocks substantially better
potential of the limited organic data and enables significant gains over traditional repetition.

\begin{table}[ht]
  \centering
  \renewcommand{\arraystretch}{1.3}
  \setlength{\tabcolsep}{3pt}
  \resizebox{1.0\columnwidth}{!}{%
    \begin{tabular}{l|cccc|cccc}
      \toprule
      & \multicolumn{4}{c|}{\textbf{400M Model (Per 0.8B Organic Tokens)}} &
      \multicolumn{4}{c}{\textbf{1.1B Model (Per 2.2B Organic Tokens)}} \\
      \textbf{Method} & \textbf{Synthesis} & \textbf{Selection} & \textbf{Pretraining} &
      \textbf{Total} & \textbf{Synthesis} & \textbf{Selection} & \textbf{Pretraining} &
      \textbf{Total} \\
      \hline
      Repeat & -- & -- & 2.7 & 2.7 & -- & -- & 19.2 & 19.2 \\
      QBSR & -- & 0.2 & 2.7 & 2.9 & -- & 0.6 & 19.2 & 19.8  \\
      MATES & -- & 0.3 & 2.7 & 3.0 & -- & 1.2 & 19.2 & 20.4 \\
      RePro & 7.8 & -- & 2.7 & 10.5 & 21.3 & -- & 19.2 & 40.5 \\
      \textsc{\ours} & 10.9 & -- & 2.7 & 13.6 & 29.8 & -- & 19.2 & 49.0 \\
      \bottomrule
    \end{tabular}
  }
  \caption{H100 GPU hours breakdown by method. Hours are computed per unit of organic
  data (0.8B tokens for 400M, 2.2B tokens for 1.1B).}
  \label{tab:compute}
\end{table}

\begin{table}[ht]
  \centering
  \renewcommand{\arraystretch}{1.3}
  \setlength{\tabcolsep}{6pt}
  \resizebox{1.0\columnwidth}{!}{%
    \begin{tabular}{l|ccc|c}
      \toprule
      \textbf{Setup} & \textbf{Generator Training} & \textbf{Generator Inference} &
      \textbf{Pretraining} & \textbf{Total} \\
      \midrule
      400M/$\alpha{=}10\%$ & 26 (11.8\%) & 33 (14.9\%) &
      162 (73.3\%) & 221 (100\%) \\
      1.1B/$\alpha{=}10\%$ & 29 (4.7\%) & 89 (14.5\%) &
      496 (80.8\%) & 614 (100\%) \\
      \bottomrule
    \end{tabular}
  }
  \caption{Decomposition of the total H100 GPU hours of \textsc{\ours}, summed over all
  three iterations. Percentages denote the share of the total cost.}
  \label{tab:cost-breakdown}
\end{table}

\subsection{Ablation of models in the generation pipeline}
\label{sec:app-pipeline-ablation}

\begin{table}[t]
  \centering
  \renewcommand{\arraystretch}{1.3}
  \setlength{\tabcolsep}{3pt}
  \resizebox{1.0\textwidth}{!}{%
    \begin{tabular}{l|c|cccc|cc|ccc|cc}
      \toprule
      & \textbf{\#Effective} &
      \multicolumn{4}{c|}{\textit{Commonsense Reasoning}} &
      \multicolumn{2}{c|}{\textit{Language Understanding}} &
      \multicolumn{3}{c|}{\textit{World Knowledge}} & & \textbf{Rec.} \\
      \textbf{Method} & \textbf{Tokens} &
      \textbf{CSQA} & \textbf{OBQA} & \textbf{PIQA} & \textbf{SIQA} &
      \textbf{HellaSwag} & \textbf{WinoG} &
      \textbf{ARC-e} & \textbf{ARC-c} & \textbf{SciQ} &
      \textbf{Avg} & \textbf{Ratio} \\
      \midrule
      \rowcolor{blue!10}
      \textsc{\ours}       &
      \textbf{34.1B}$_{~\textcolor{red}{\textbf{5.2}\times}}$ & \underline{0.3882} &
      \textbf{0.3420} &
      \textbf{0.6649} & \underline{0.4284} & \underline{0.3868} & 0.5201 &
      \underline{0.6123} &
      \textbf{0.3278} & \textbf{0.8540} & \textbf{0.5027} & \textbf{79\%} \\
      ~~w/ Qwen3-Embedding     & \underline{33.8B}$_{~\textcolor{red}{5.1\times}}$ &
      \textbf{0.3939} & 0.3360 & \underline{0.6610} & 0.4278 &
      \textbf{0.3941} & \underline{0.5264} & 0.6070 & \underline{0.3211} &
      \underline{0.8480} & \underline{0.5017} & \underline{77\%} \\
      ~~w/ Self-Generation            & 32.0B$_{~\textcolor{red}{4.8\times}}$ &
      0.3776 & \underline{0.3400} & 0.6420 & \textbf{0.4289} &
      0.3726 & \textbf{0.5304} & \textbf{0.6404} & 0.3043 & 0.8450 &
      0.4979 & 68\% \\
      \hline
      Repeat                          & 6.6B$_{~{1.0\times}}$ & 0.3358 &
      0.3140 & 0.6442 & 0.4135 & 0.3519 & \textbf{0.5304} & 0.5579 & 0.2676 & 0.7920 &
      0.4675 & 0\% \\
      \bottomrule
    \end{tabular}
  }
  \caption{Ablation of the models adopted in our generation pipeline, on the
    400M/$\alpha{=}10\%$ setting.
    ``w/ Qwen3-Embedding'' replaces BERTScore with Qwen3-Embedding-0.6B
    similarity as the semantic faithfulness reward, and ``w/ Self-Generation''
    initializes the generator from our own pretrained checkpoint instead of
  OLMo2-1B-Instruct.}
  \label{tab:pipeline-ablation}
\end{table}

Our generation pipeline involves two key models, BERTScore~\citep{zhang2020bertscore}
as the semantic faithfulness reward and OLMo2-1B-Instruct as the initialization of the
generator.
We ablate both choices in Table~\ref{tab:pipeline-ablation}.

Replacing BERTScore with the embedding similarity from
Qwen3-Embedding-0.6B~\citep{zhang2025qwen3embedding} yields nearly identical
performance (0.5017 vs.\ \textbf{0.5027}), indicating that our faithfulness reward is
robust to the choice of embedding backbone.

We further train the generator from our own pretrained model, by
instruction-tuning the 400M checkpoint of the Repeat baseline on
FLAN~\citep{wei2022flan} and applying the same RL pipeline, which forms a closed-loop
``self-generation'' setup.
This variant still substantially outperforms Repeat (\textbf{0.4979} vs.\ 0.4675) and
approaches the generator initialized from OLMo2, confirming that the gains
of \textsc{\ours} still hold without an external generator.

\subsection{High quality of generated synthetic data}
\label{sec:app-quality}

We evaluate the quality of generated synthetic data along two dimensions: intrinsic
text quality via DataMan~\citep{peng2025dataman} scores and benefits to the
pretraining model via data influence scores.
We use the same 1{,}000 sampled documents as in \S\ref{sec:results-faithfulness} and
compare three conditions: organic data, the base generator, and \textsc{\ours}
(using the $\pi_1$ checkpoint from the 400M setting at 21.8B training tokens).
Influence scores are computed following Eq.~\ref{eq:influence}, using pretraining
checkpoints at 21.8B training tokens as well.

\begin{figure*}[t]
  \centering
  \begin{subfigure}[t]
    {0.2365\textwidth}
    \raisebox{0pt}{\includegraphics[width=1.0\linewidth]{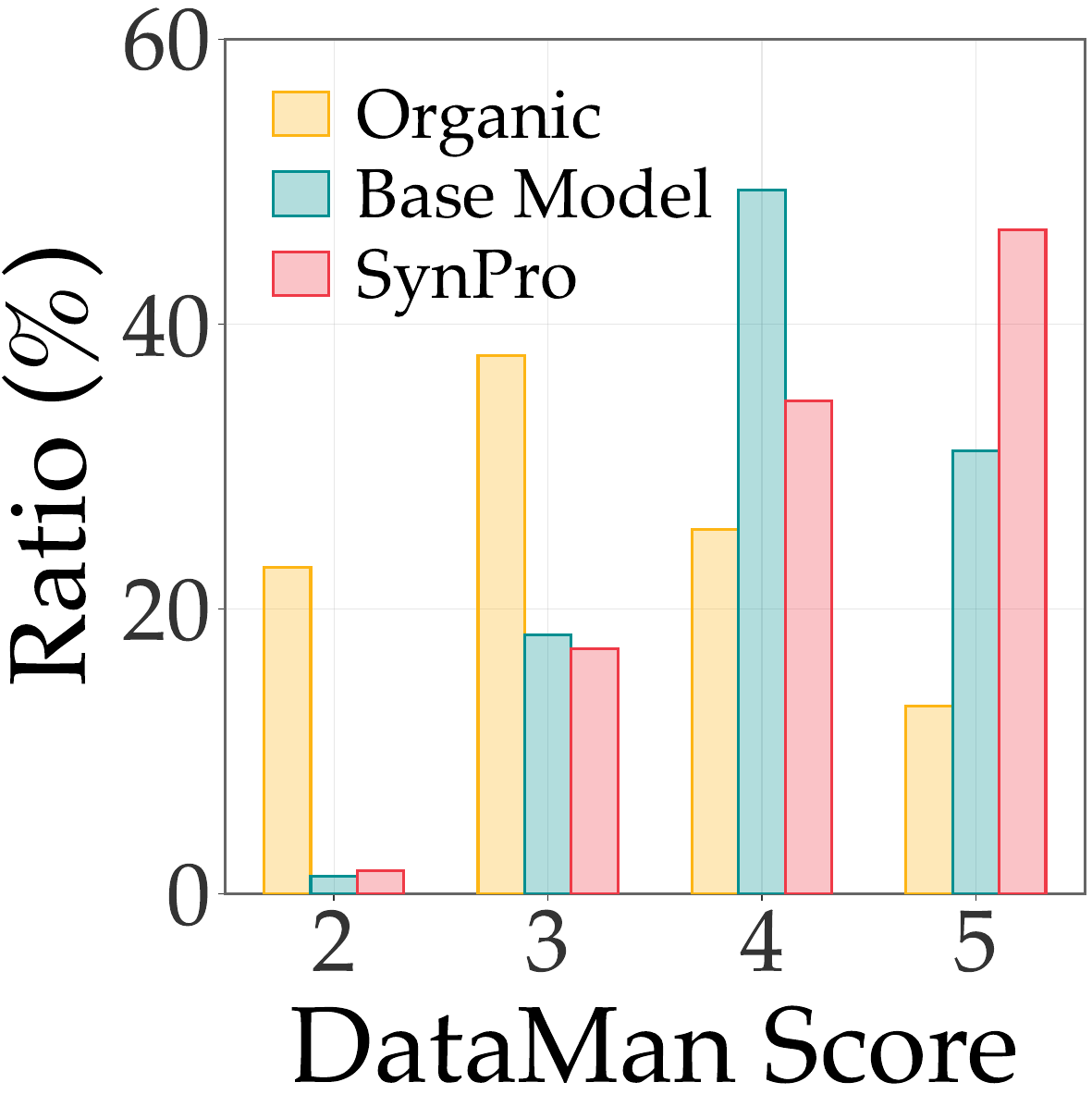}}
    \caption{Rephrase:DataMan}
    \label{fig:quality-a}
  \end{subfigure}
  ~
  \begin{subfigure}[t]
    {0.2365\textwidth}
    \centering
    \raisebox{0pt}{\includegraphics[width=1.0\linewidth]{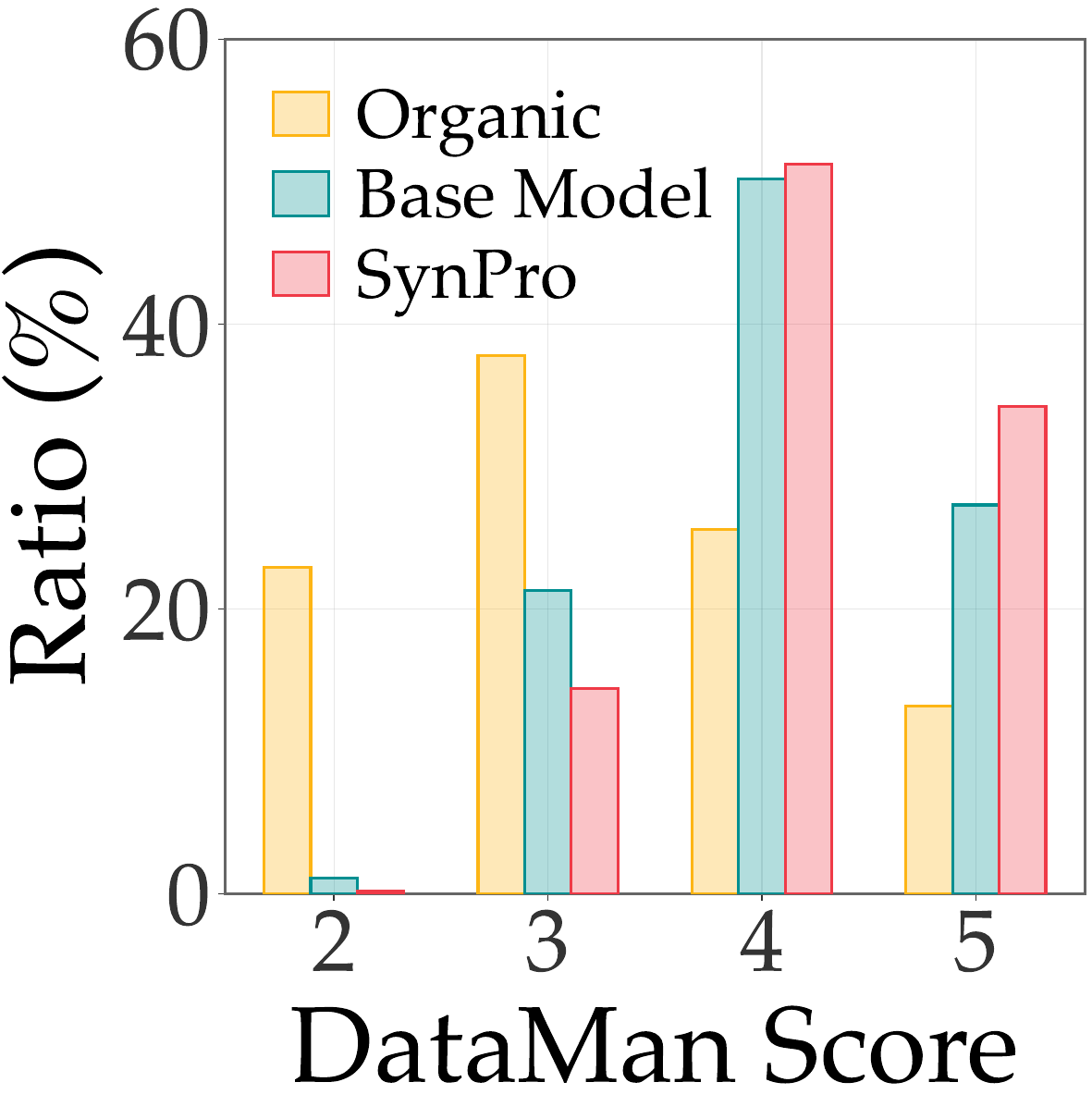}}
    \caption{Reformat:DataMan}
    \label{fig:quality-c}
  \end{subfigure}
  ~
  \begin{subfigure}[t]
    {0.2375\textwidth}
    \centering
    \raisebox{0pt}{\includegraphics[width=1.0\linewidth]{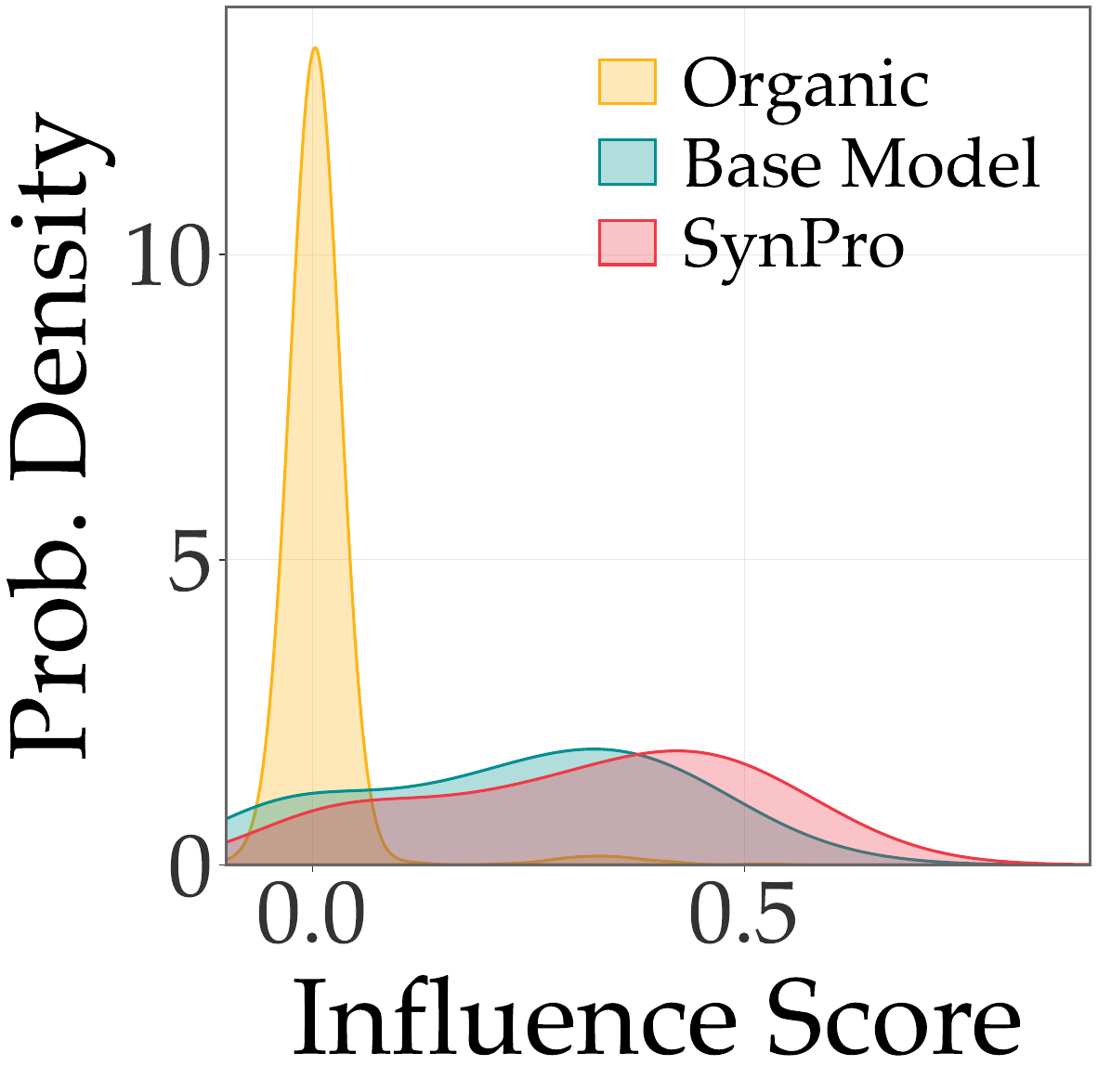}}
    \caption{Rephrase:Influence}
    \label{fig:quality-b}
  \end{subfigure}
  ~
  \begin{subfigure}[t]
    {0.2375\textwidth}
    \centering
    \raisebox{0pt}{\includegraphics[width=1.0\linewidth]{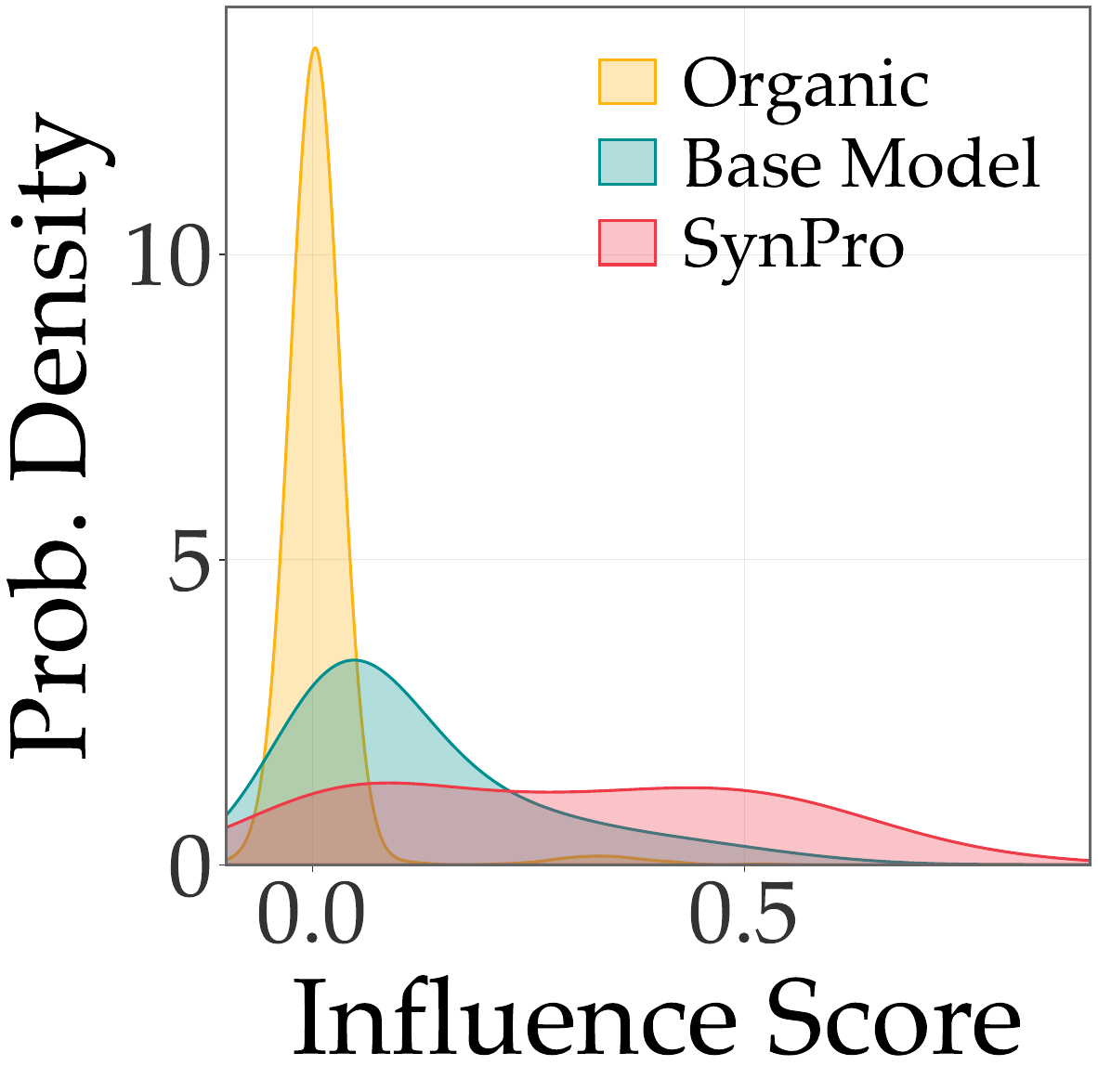}}
    \caption{Reformat:Influence}
    \label{fig:quality-d}
  \end{subfigure}
  \caption{Quality analysis on 1{,}000 randomly sampled organic documents not seen in RL.
    (a,\,b)~DataMan score and (c,\,d)~data influence score
    for rephrasing and reformatting, respectively. \textsc{\ours} uses $\pi_1$ checkpoint from
  the 400M setting at 22B training tokens.}
  \label{fig:quality}
  \vspace{-0.2cm}
\end{figure*}

\paragraph{DataMan score.}
Figures~\ref{fig:quality-a} and~\ref{fig:quality-c} show the DataMan score
distributions for rephrasing and reformatting, respectively.
\textsc{\ours} substantially improves over the organic data (mean 3.28), which
achieves a mean score of \textbf{4.26} for rephrasing and
\textbf{4.19} for reformatting.
In comparison, the base generator achieves a mean score of 4.10 for rephrasing and 4.04
for reformatting.
The shift toward higher scores (particularly 5) confirms that our quality
reward drives the generator to produce more coherent and well-structured text while
maintaining faithfulness.

\paragraph{Data influence.}
Figures~\ref{fig:quality-b} and~\ref{fig:quality-d} show the influence score
distributions.
Organic data has near-zero mean influence (0.01), reflecting that the pretraining
model has already absorbed most of organic content through repeated exposure.
For rephrasing, \textsc{\ours} achieves a mean influence of \textbf{0.32}, a moderate
improvement compared
to 0.28 from the base generator since rephrasings largely
preserve the original content.
The effect is more pronounced for reformatting, where \textsc{\ours} reaches a mean of
\textbf{0.29}, doubling the base generator's 0.12.
This gap arises as the reformatting operation produces structurally novel outputs
(e.g., QA pairs, reasoning traces) that present familiar content in forms the model
has not yet seen, and the influence reward further steers generation toward content
the model finds most informative.
We also find the influence score to be nearly uncorrelated with the semantic
faithfulness score on these documents (Pearson $-0.09$, Spearman $-0.09$), indicating
that the two rewards capture complementary rather than redundant signals.

\subsection{RL training dynamics}
\label{sec:app-rl-dynamics}

\begin{figure}[t]
  \vspace{-0.2cm}
  \centering
  \begin{subfigure}[t]{0.35\textwidth}
    \centering
    \includegraphics[width=1.0\linewidth]{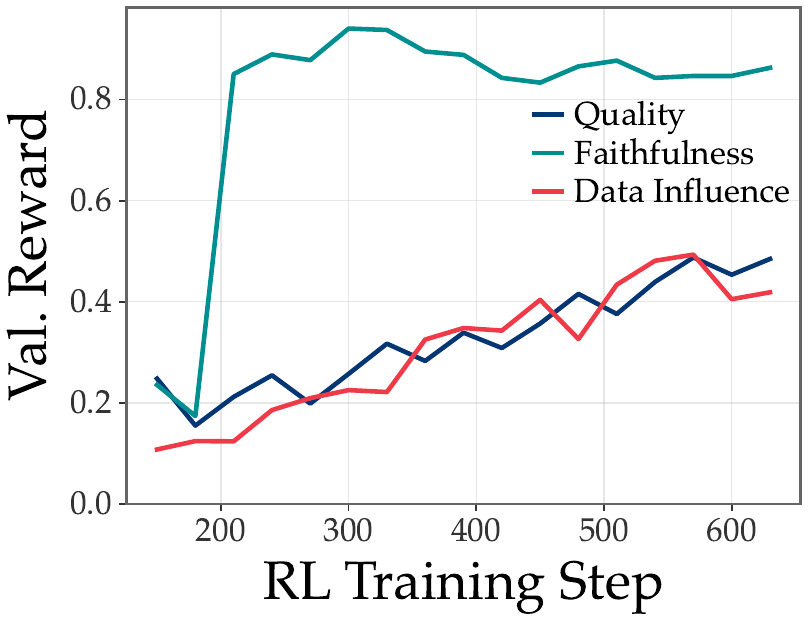}
    \caption{Rephrasing reward curves}
    \label{fig:rephrase-curves}
  \end{subfigure}
  \begin{subfigure}[t]{0.35\textwidth}
    \centering
    \includegraphics[width=1.0\linewidth]{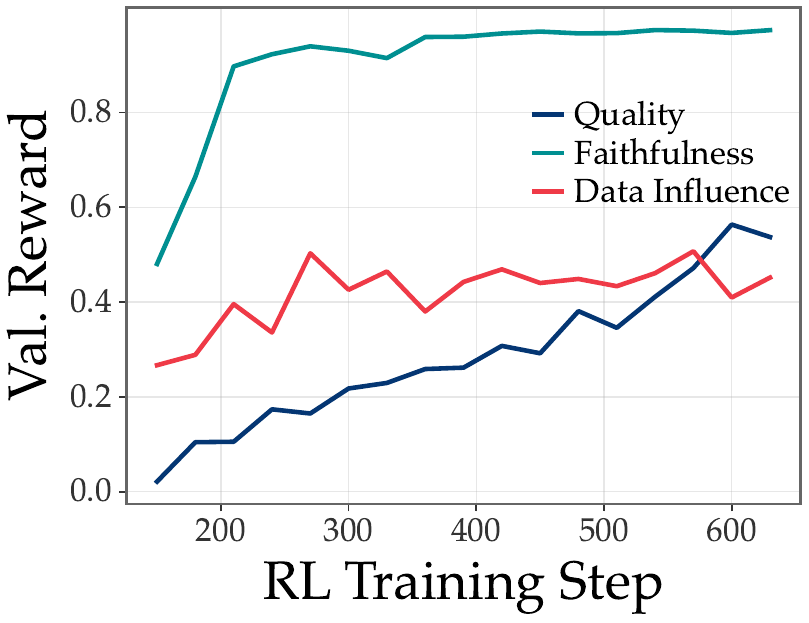}
    \caption{Reformat reward curves}
    \label{fig:reformat-curves}
  \end{subfigure}
  \caption{Example validation reward (400M pretraining model, $\pi_1$ generator) curves
  during RL training for (a) rephrasing and (b) reformatting.}
  \vspace{-0.3cm}
  \label{fig:rl-training-curves}
\end{figure}

Figure~\ref{fig:rl-training-curves} shows example validation reward curves during RL
training for the $\pi_1$ generator in the 400M setting. A first observation is that the
three rewards
can be improved jointly rather than trading off sharply against one another. For both
rephrasing and reformatting, the quality, faithfulness, and data influence rewards all
rise substantially over training and remain high near convergence, indicating that the
generator can simultaneously become more coherent, more source-grounded, and more
useful to the current pretraining model.

Second, the faithfulness reward increases especially quickly in the early stage of
training and reaches a high level well before convergence. Intuitively,
faithfulness is a relatively easy signal to optimize early on, as the generator quickly
learns to stay close to the source document and avoid unsupported generations,
after which later training focuses more on improving quality and informativeness.

Finally, our reward coefficients achieve a balance among these objectives. As described
in \S\ref{sec:setup}, we set $\lambda_{\text{quality}} = 1$,
$\lambda_{\text{faithful}} = 1$, and $\lambda_{\text{influence}} = 3$, so that the
combined contribution of the quality and data influence rewards is on a similar scale
to the faithfulness reward. This setting prevents the policy from over-optimizing
faithfulness alone, achieving a practical balance
between faithfulness and informative synthetic data generation.
We tuned these coefficients by validation trends, but this is a common one-time overhead in RL.
For other RL hyperparameters, we follow best practices in RePro~\citep{yu2025repro}.

\section{Case study}
\label{sec:app-cases}

In this section, we present representative examples showing how \textsc{\ours}
transforms organic data into faithful and model-aware synthetic data.

The rephrasing comparison highlights the effect of the faithfulness reward more
directly. Relative to the rephrasing of base model (OLMo2-1B-Instruct), both $\pi_0$ and
$\pi_1$ stay much
closer to the source content and preserve the original instructional structure,
instead of drifting into generic advice or summary-style text. Among them, $\pi_1$
is more factually abundant than $\pi_0$, consistent with the
additional effect of the data influence reward.

We next turn to reformatting cases, where the blue spans mark source-grounded evidence
that directly supports the reformatted outputs.
In both examples, all shown outputs remain grounded in the source, but the
initial policy $\pi_0$ tends to produce broader and more surface-level reformats,
while later model-aware policies place greater emphasis on advanced factual content that
helps the model improve.
The intermediate policy $\pi_1$ already moves in this direction, but $\pi_2$ makes the
shift more pronounced.

Specifically, in Case~1, $\pi_0$ mostly asks local identification questions (e.g., ``saturation,''
``gaze confirmation,'' and ``Fitts Law''), whereas $\pi_2$ shifts toward more concrete
takesaways about VR stress, trauma, and ethical concerns. In Case~2, $\pi_0$ already
extracts several factual items, but they remain relatively shallow and loosely
organized. $\pi_1$ increases specificity, while $\pi_2$ further concentrates on
explicit factual content about the Delphi method, including its
origin, purpose, and questionnaire-based aggregation. Importantly, Case~2 contains two
largely independent pieces of information, one about the Delphi method and one about
the militia system, yet the generator is able to reformat salient content from both
sources rather than collapsing onto only one topic. Together, these cases strongly support the
trend in Figure~\ref{fig:awareness-d}, where the generator shifts away from broader,
surface-level reformats and toward more factual, source-grounded outputs across policy
updates.

\setlength{\columnseprule}{0.7pt}
\def\columnseprulecolor{\color{gray!70}}

\begin{tcolorbox}[
    colback=blue!5!white,
    colframe=blue!75!black,
    breakable,
    title=\textsc{\ours}: Rephrasing Case
  ]

  \textbf{Organic (truncated)} \\[4pt]
  you want to get into the action and show up during my ACT. But you need to leave
  for a while. We can talk later.'' By naming the monster, you can help tame the
  monster. Creating some cognitive distance from the anxious thoughts allows you to
  achieve a measure of control over them.
  13. Imagined practice makes perfect: If you have had many experiences of anxiety
  during tests, it may help to visualize yourself taking a test without suffering the
  effects of anxiety. It's a practice known as cognitive rehearsal, or guided
  imagery, taking a mental walk-through in advance of a performance event. Athletes
  do it all the time, imagining themselves performing at their peak level, in
  advance of the high-stakes event. Imagine yourself walking through the test,
  missing problems and staying calm and centered: Lay down a new template of you as
  a peak-tester, and make the images as vivid and sensorily rich as possible, so
  your mind believes them. Imagined practice can be as powerful as actual practice.
  14. Regulate your breathing: Deep diaphragmatic breathing is a powerful
  anxiety-reducing technique because it activates the body's relaxation response.
  Breathing from the diaphragm, in a slow measured way, filling the stomach then the
  chest, stimulates the vagus nerve, which controls the parasympathetic (calming)
  nervous system. Stimulating the vagus nerve leads to a decrease in heart rate,
  blood pressure and other sympathetic responses. Practice breathing deeply and
  consciously, multiple times per day, and this technique will be available to you
  on test day.
  15. Imagine you are breathing into your heart center: There's a simple technique
  I use when I'm nervous, where I place my hand over my heart, and take deep
  breaths, focusing my energy on my heart. This is a technique I learned from an
  organization called Heartmath. The act of combining slow breathing, a nurturing
  gesture, and redirecting my consciousness to my heart, helps restore a sense of
  calm.

  \vspace{6pt}
  \hrule
  \vspace{6pt}

  \textbf{OLMo2-1B-Instruct Rephrasing (truncated)} \\[4pt]
  In order to take part in the ACT, you must first excuse yourself temporarily.
  Discussing it later, however, will be beneficial. Through naming the monster, one
  can control it effectively. Preparing mentally for the exam without being
  overwhelmed by anxiety is similar to imagining yourself performing well and
  staying composed during the test. Athletes regularly employ such practices,
  visualizing themselves succeeding. This mental rehearsal can be just as effective
  as actual study. Here is a paraphrased version: Maintain a balance in your anxiety
  before the ACT, by stepping away and discussing later. Dealing with anxiety
  mentally can make you more in control of it. By acknowledging the presence of the
  monster, you're more likely to manage it. Imagining yourself as calm and
  successful during the test can prepare you effectively. Regular exercises like
  deep breathing, focusing on your heart, visualizing in a safe way, regular
  physical activity, ensuring you sleep well, practicing relaxation techniques, and
  engaging in conversations or even meditating can all help reduce anxiety. These
  strategies are applicable across various aspects of life and education,
  emphasizing the importance of self-regulation. It's important to note that the
  listed techniques cover various facets of mental health management, from physical
  relaxation to psychological coping strategies, and their general applicability
  extends far beyond the context of an ACT test. The goal is to build resilience and
  emotional regulation abilities which are universally useful.

  \vspace{6pt}
  \hrule
  \vspace{6pt}

  \textbf{$\pi_0$ Rephrasing (truncated)} \\[4pt]
  You want to engage actively and participate during my ACT. However, you must
  temporarily disconnect for discussion. We can discuss later. By labeling the
  monster, you can help mitigate its effects. Creating some mental distance from
  anxious thoughts assists in gaining control over them. 13. Imagined rehearsal
  brings perfection practice: If you have frequently experienced anxiety during
  exams, envisioning yourself performing without experiencing anxiety-induced
  discomfort can serve as a practice known as cognitive rehearsal or guided
  imagery. Athletes regularly engage in this technique by visualizing themselves at
  their peak levels before performance events. Imagine walking through the exam,
  missing questions and maintaining composure and centering: Develop a new template
  of yourself as a high-performing test-taker, enhancing your mental imagery vividly
  and sensorially. 14. Control breathing: Deep diaphragmatic breathing is a potent
  anxiety-reducing method due to its stimulation of the body's relaxation response.
  Breathing from the diaphragm, in a measured slow manner, fills the stomach then
  the chest, activates the vagus nerve, and reduces heart rate, blood pressure, and
  other sympathetic reactions. Practicing deep breathing multiple times per day and
  employing this technique during test days will be beneficial. 15. Visualize
  breathing into your heart: I employ a simple technique when feeling anxious where
  I place my hand over my heart and engage in slow breathing, focusing my
  consciousness on my heart. This technique was learned from an organization called
  Heartmath. The action of combining slow breathing, a nurturing gesture, and
  redirecting mental focus to your heart, helps restore a sense of calm.

  \vspace{6pt}
  \hrule
  \vspace{6pt}

  \textbf{$\pi_1$ Rephrasing (truncated)} \\[4pt]
  you might want to dive into action and showcase up during my ACT, but you need
  to take a break beforehand. We can discuss this further later. By naming the
  monster, you can help calm it down. Creating some cognitive distance from anxious
  thoughts allows you to gain a sense of control over them. 13. Imagined practice
  makes perfect: If you have experienced many instances of anxiety during tests,
  visualize yourself taking a test without letting anxiety interfere. This is known
  as cognitive rehearsal, or guided imagery, taking a mental walk-through before a
  performance. Athletes do this all the time, imagining themselves performing at
  their best, in advance of a crucial event. Visualize yourself walking through the
  test, bypassing difficult problems and staying calm and centered: Lay down a new
  template of yourself as a peak-tester, and fill it with as many sensory details
  as possible so your mind believes it. Imagined practice can be as effective as
  actual practice. 14. regulate your breathing: Deep, diaphragmatic breathing is a
  powerful technique for reducing anxiety, as it activates the body's relaxation
  response. Breathing from the diaphragm, in a slow and measured way, filling the
  stomach then the chest, stimulates the vagus nerve, which controls the
  parasympathetic (calming) nervous system. Stimulation of the vagus nerve reduces
  heart rate, lowers blood pressure, and diminishes other sympathetic responses.
  Practice breathing deeply and consciously, multiple times a day, and this
  technique will be ready for you on test day. 15. Imagine breathing into your
  heart center: I have a simple technique I use when feeling nervous, where I place
  my hand on my heart and take deep breaths, focusing my energy on my heart. This
  technique, which I learned from an organization called Heartmath, combines slow
  breathing, a nurturing gesture, and redirecting my consciousness to my heart,
  helps restore a sense of calm.

\end{tcolorbox}

\vspace{8pt}

\begin{tcolorbox}[
    colback=blue!5!white,
    colframe=blue!75!black,
    title=\textsc{\ours}: Reformat Case 1
  ]
  \begin{multicols}{2}

    \textbf{Organic (truncated)} \\[4pt]
    The more saturated an object or image compared to its surroundings, the more likely
    it is to \textcolor{blue!70!black}{draw attention}. \dots Again,
    \textcolor{blue!70!black}{saturation} can be used in
    VR to \textcolor{blue!70!black}{direct user's attention} in the same way it has been
    used in media previously. \dots This author suggests using
    \textcolor{blue!70!black}{desaturation}
    to \textcolor{blue!70!black}{keep user's focused in 360-degree VR environments}. \dots
    In VR, \textcolor{blue!70!black}{gaze input} is becoming common
    practice. Often, a means
    of \textcolor{blue!70!black}{confirmation} is to have the user gaze at an object or
    image for a \textcolor{blue!70!black}{set amount of time} to confirm
    \textcolor{blue!70!black}{critical actions}. \dots Most Real-Time
    Strategy games use a \textcolor{blue!70!black}{positive
    feedback loop} by
    encouraging the player to gather resources to build factories or other structures.
    \textcolor{blue!70!black}{These new structures then produce even more resources}. \dots
    \textcolor{blue!70!black}{The time required to
    move to a target is a function of the target size and distance to the target}
    (Lidwell, 2010). \dots
    In VR \textcolor{blue!70!black}{Fitts law is still relevant}. \dots VR experiences have
    the same opportunity to induce stress in the user. \dots A user in VR is completely
    immersed in the experience, and \textcolor{blue!70!black}{traumatic events in VR can seem real}
    to the user.

    \vspace{6pt}
    \hrule
    \vspace{6pt}

    \textbf{$\pi_1$ Reformat (truncated)} \\[4pt]
    - Question: What is mentioned as an advantage of using desaturation?
    Answer: To keep user's focused in 360-degree VR environments. \\
    - Question: How does VR confirmation work?
    Answer: Confirmation is used to confirm ``critical actions, inputs, or
    commands.'' \\
    - Question: What caution is given regarding VR experiences?
    Answer: A designer must question the ethics of creating obscenely graphic,
    disturbing, and horrifying VR experiences. \\

    \columnbreak

    \textbf{$\pi_0$ Reformat (truncated)} \\[4pt]
    - Question: What design principle is discussed that involves directing user
    attention? Answer: Saturation \\
    - Question: What type of interaction does the text discuss that relies on user
    gaze input? Answer: Gaze confirmation \\
    - Question: How does feedback in video games positively affect continued play?
    Answer: Encourage player gathering resources and expanding structures \\
    - Question: What law was discussed regarding mouse movement in computing?
    Answer: Fitts Law \\
    - Question: What does this law predict? Answer: The time it takes for an
    individual to move a cursor from one point to another \\
    - Question: VR experiences can induce a similar stress response. What might this
    include? Answer: Traumatic events

    \vspace{6pt}
    \hrule
    \vspace{6pt}

    \textbf{$\pi_2$ Reformat (truncated)} \\[4pt]
    - Question: What does the text discuss? \\
    - Answer: Fitts Law principle in VR as described through anecdotal
    examples. \\
    - Question: What factor does the text emphasize about immersive VR experiences? \\
    - Answer: Extreme immersion can induce severe stress and potential
    lasting mental effects. \\
    - Question: What form of stress is mentioned in the text? \\
    - Answer: Stress induced through intense, realistic VR environments
    that can create psychologically impactful experiences. \\
    - Question: What consideration for VR interaction is highlighted? \\
    - Answer: The importance of ethical considerations for content
    designers due to the heightened susceptibility to emotional and psychological
    effects of VR experiences. \\

  \end{multicols}
\end{tcolorbox}

\vspace{8pt}

\begin{tcolorbox}[
    colback=blue!5!white,
    colframe=blue!75!black,
    title=\textsc{\ours}: Reformat Case 2
  ]
  \begin{multicols}{2}

    \textbf{Organic (truncated)} \\[4pt]
    The \textcolor{blue!70!black}{Delphi method} was developed in the
    \textcolor{blue!70!black}{United States}
    in the \textcolor{blue!70!black}{1950s and 1960s} in
    the \textcolor{blue!70!black}{military domain}. It has been considered particularly
    useful in helping
    researchers \textcolor{blue!70!black}{determine the range of opinions} which exist on
    a particular subject,
    in investigating issues of policy or clinical relevance and in trying to come
    to a \textcolor{blue!70!black}{consensus on controversial issues}. The objectives can
    be roughly divided
    into those which aim to \textcolor{blue!70!black}{measure diversity} and those which aim to
    \textcolor{blue!70!black}{reach consensus}. \dots Different ways to employ this method
    have been devised \dots comment on a \textcolor{blue!70!black}{questionnaire} \dots \dots
    The \textcolor{blue!70!black}{militia system} was treasured by many people because
    the existence of a
    well-regulated militia, composed of civilians readily available for emergency
    military service, tended to deprive the government of an excuse for maintaining
    \textcolor{blue!70!black}{standing armies}. \dots
    \textcolor{blue!70!black}{Alexander Hamilton}, for one, complained that the militia
    system violated the
    economic principle of division of labor. \dots Citizens were always going to
    resist \textcolor{blue!70!black}{unpaid military training} \dots

    \vspace{6pt}
    \hrule
    \vspace{6pt}

    \textbf{$\pi_1$ Reformat (truncated)} \\[4pt]
    - Question: What is the Delphi method? Answer: Developed in the United States
    in the 1950s and 1960s in the military domain. \\
    - Question: What is its primary use? Answer: Determining the range of opinions
    on a particular subject, investigating issues of policy or clinical relevance,
    and coming to a consensus on controversial issues. \\
    - Question: What aim does the Delphi method have? Answer: Measure diversity and
    consensus on contentious subjects. \\
    - Question: Who complained that the militia system violated economic principles?
    Answer: Alexander Hamilton. \\
    - Question: What was the issue identified as being problematic about the militia
    system? Answer: The militia system was noted as being ineffective due to citizen
    resistance to unpaid military service. \\

    \columnbreak

    \textbf{$\pi_0$ Reformat (truncated)} \\[4pt]
    - Question: What kind of a method was developed in the US in the 50s and 60s?
    Answer: Delphi method \\
    - Question: In what domain did the Delphi method emerge? Answer: Military \\
    - Question: What was the method meant to do? Answer: Measure diversity \\
    - Question: What system was criticized for depriving the government of an excuse
    to maintain a standing army? Answer: The militia system \\
    - Question: Why did some comment about the militia system? Answer: Because it
    served the interests of a government needing less risk of using permanent armies. \\
    - Question: What concerns were voiced regarding the militia? Answer: The
    necessity of citizens resisting unpaid military service \\

    \vspace{6pt}
    \hrule
    \vspace{6pt}

    \textbf{$\pi_2$ Reformat (truncated)} \\[4pt]
    - Question: What is the origin of the Delphi method? \\
    - Answer: Military in the United States in the 1950s and 1960s \\
    - Question: What is the purpose of the Delphi method? \\
    - Answer: Measuring diversity and developing consensus \\
    - Question: How are experts' opinions aggregated in Delphi method sessions? \\
    - Answer: Through structured questionnaires \\
    - Question: How was the risk of federal oppression addressed by the drafters of
    the Constitution? \\
    - Answer: The problem of oppression is mitigated by allowing a standing army with
    the right of Congress to call forth the militia. \\
    - Question: Why might the government prefer to maintain a well-trained militia
    rather than a standing army? \\
    - Answer: A militia that is disciplined and reliable is more effective in
    protecting liberty than an army that might turn against its citizens. \\

  \end{multicols}
\end{tcolorbox}

\section{Prompts}
\label{sec:app-prompts}

This section provides the detailed prompts used for each evaluation in our paper. We
adapt and modify prompts from prior works such as RePro~\citep{yu2025repro} and
Nemotron-CC~\citep{su2025nemotroncc}, and also design new prompts for specific evaluations. The
prompts are designed to be clear, specific, and aligned with the goal we want to achieve.

\setlength{\fboxsep}{1pt}
\newcommand{\mybox}[2]{\colorbox{#1!20}{#2}}

\begin{tcolorbox}[colback=brown!8, colframe=brown!50!black, title=Rephrasing Prompt
  (adapted from RePro)]
  Your task is to read and paraphrase the provided text following these instructions:

  - Delete clearly irrelevant content:

  \quad- Website headers, navigation bars, or menu items (e.g., ``Home | About | Contact'')

  \quad- Unrelated HTTP links (e.g., ads, trackers, developer tools)

  \quad- Generic footers (e.g., contact info, privacy policies, unsubscribe links)

  \quad- Empty lines or decorative elements (e.g., ``---'')

  - Preserve all content that is relevant and meaningful:

  \quad- Informative or independently useful

  \quad- Related to the topic, even tangentially

  \quad- Provides context, background, or supporting value

  \quad- Includes technical terms, key concepts, factual details, reasoning, and examples

  - Handle mixed-relevance sentences carefully:

  \quad- Remove only the irrelevant fragment if the rest remains coherent

  \quad- Delete the whole sentence if the remainder loses meaning

  - Do not alter meaningful content unnecessarily:

  \quad- Only delete or modify when content is clearly meaningless or off-topic

  \quad- Preserve the original structure, logic, and depth of the text

  - Do not add explanations, notes, assumptions, or claims not found in the original text

  Here is the text:

  \mybox{blue}{\{Organic Text\}}

  Task:

  After thoroughly reading the above text, paraphrase it in high-quality and clear
  English following the instructions.

  Start your response immediately with ``Here is a paraphrased version:'' and then
  provide the paraphrased text.
\end{tcolorbox}

\begin{tcolorbox}[colback=brown!8, colframe=brown!50!black, title=Reformat Prompt (adapted from
  Nemotron-CC)]
  Task: Read the text and convert it into a different format.

  Follow these instructions:

  1. Ask diverse questions that require different cognitive skills or cover different
  aspects of the text.

  2. Ask questions in various forms such as:

  \quad- Yes/No questions that require determining whether a statement is true or false.

  \quad- Open-ended questions that begin with words like what, how, when, where, why and who.

  \quad- Multi-choice questions that offers two or more options to choose from. Include
  the options in the question.

  \quad- Comparison questions that compare two quantities or objects and determine the
  relationship between them.

  \quad- Reading comprehension questions that test the ability to understand and analyze the text.

  \quad- Problem-solving questions that test the ability to solve mathematical, physical,
  or logical problems.

  3. Focus on asking questions about factual information, important knowledge, or
  concrete details in the text.

  4. Write questions and answers using clear and concise language.

  5. Use plain text. Do not use Markdown.

  6. Each question and answer pair should be on a separate line. Tag the question with
  ``Question:'' and the answer with ``Answer:''.

  Text:

  \mybox{blue}{\{Organic Text\}}

  Task:

  After reading the above text, ask up to 8 questions and provide the correct answers
  following the instructions. Give your response in this format:

  Here are the questions and answers based on the provided text:

  - Question: [first question] Answer: [first answer]

  - Question: [second question] Answer: [second answer]

  ....
\end{tcolorbox}

\begin{tcolorbox}[colback=brown!8, colframe=brown!50!black, title=DataMan Prompt]
  Please score the text on fourteen evaluation criteria and specify its domain:

  Text: \mybox{blue}{\{Text\}}

  Domain:\_

  [1]Accuracy:\_/5

  [2]Coherence:\_/5

  [3]Language Consistency:\_/5

  [4]Semantic Density:\_/5

  [5]Knowledge Novelty:\_/5

  [6]Topic Focus:\_/5

  [7]Creativity:\_/5

  [8]Professionalism:\_/5

  [9]Style Consistency:\_/5

  [10]Grammatical Diversity:\_/5

  [11]Structural Standardization:\_/5

  [12]Originality:\_/5

  [13]Sensitivity:\_/5

  [14]Overall Score:\_/5
\end{tcolorbox}

\begin{tcolorbox}[colback=brown!8, colframe=brown!50!black, title=Structural Faithfulness
  Prompt (adapted from RePro)]
  [Instruction]

  You are given two pieces of text: an original pretraining data sample and a rephrased version.

  Your task is to judge if the rephrased version preserves the **structure** of the original sample.

  - By ``structure'', we mean formatting, style, and presentation (e.g., paragraphing,
  JSON, list format, code blocks, markdown usage, plain text style).

  - Do NOT consider semantic meaning. Ignore whether the words are the same or the
  content is equivalent.

  - Focus only on whether the rephrased sample follows the same textual structure as the
  original (e.g., if the original is plain text paragraphs, the rephrased should also be
  plain text; if the original has bullet lists, the rephrased should also have bullet lists).

  [Output]

  Output **only** `1` if the structure is preserved.

  Output **only** `0` if the structure is not preserved.

  [Examples]

  Example 1:

  Original:

  This is a paragraph.

  This is another line.

  Rephrased:

  Here is a rewritten paragraph.

  Here is another line of text.

  Explanation: Both are plain text paragraphs, no special formatting. Structure preserved. Output: 1

  ---

  Example 2:

  Original:

  - Item one

  - Item two

  Rephrased:

  First item. Second item.

  Explanation: The original uses a bullet list, while the rephrased is plain sentences.
  Structure not preserved. Output: 0

  ---

  Example 3:

  Original:

  \{\{``name'': ``Alice'', ``age'': 30\}\}

  Rephrased:
  \{\{``person'': ``A.'', ``years'': 30\}\}

  Explanation: Both are JSON objects with the same structured format. Structure preserved. Output: 1

  Explanation: The original is plain code with no markdown fences, while the rephrased
  introduces code fences. Structure not preserved. Output: 0

  [Original]

  \mybox{blue}{\{Organic Text\}}

  [Rephrased]

  \mybox{blue}{\{Rephrased Text\}}
\end{tcolorbox}

\begin{tcolorbox}[colback=brown!8, colframe=brown!50!black, title=Reformat
  Faithfulness Prompt]
  You are given an original text passage and a list of topic-content pairs
  generated from it. For each pair, determine whether it is faithful to the
  original text.

  A faithful reformat pair satisfies ALL of the following:

  1. The topic is about a subject covered or clearly implied by the original text.

  2. The content is correct and supported by the original text.

  3. The content does not contradict the text.

  Labels:

  - ``Faithful'': the topic is relevant and the content is correct.

  - ``Unfaithful\_Topic'': the topic covers something not covered or implied by
  the text.

  - ``Unfaithful\_Content'': the topic is valid but the content is wrong or
  unsupported.

  Original Text:

  \mybox{blue}{\{Organic Text\}}

  Reformat Pairs:

  \mybox{blue}{\{Reformat Text\}}

  Respond with exactly one label per line in this format (no extra text):

  1. [label]

  2. [label]

  ...
\end{tcolorbox}

\begin{tcolorbox}[colback=brown!8, colframe=brown!50!black, title=Reformat Type
  Classification Prompt]
  Classify each question into exactly one of the following four categories:

  Categories:

  - ``Factual'': Questions asking for a specific fact such as who, what, when, where,
  or which. The answer is a concrete entity, date, number, or name directly stated in
  the text.

  - ``Analytical'': Questions asking for reasoning such as why, how, cause-effect, or
  mechanisms. The answer explains a process, reason, or causal relationship.

  - ``Conceptual'': Questions asking for definitions, explanations, or meaning such as
  what does X mean, what is X, or describe X. The answer clarifies a concept or term.

  - ``Integrative'': Questions requiring combining multiple pieces of information such
  as summarize, what is the overall takeaway, or how do X and Y relate. The answer
  synthesizes across the text.

  Text:

  \mybox{blue}{\{Reformat Text\}}

  Respond with exactly one category per line in this format (no extra text):

  1. [category]

  2. [category]

  ...
\end{tcolorbox}

\end{document}